\documentclass[lettersize,journal]{IEEEtran}
\usepackage{amsmath,amsfonts}
\usepackage{algorithmic}
\usepackage{algorithm}
\usepackage{array}

\usepackage[caption=false,font=footnotesize,labelfont=rm,textfont=rm]{subfig}
\usepackage{textcomp}
\usepackage{stfloats}
\usepackage{url}
\usepackage{verbatim}
\usepackage{graphicx}
\usepackage{booktabs}
\usepackage[normalem]{ulem}
\usepackage{subfloat}
\DeclareSubrefFormat{parens}{#1(#2)} 

\usepackage{booktabs}
\usepackage{multirow}
\usepackage[table,xcdraw]{xcolor}
\definecolor{myviolet}{RGB}{237,2,140}
\usepackage[normalem]{ulem}
\usepackage{cite}
\usepackage{adjustbox}
\usepackage{amssymb}
\begin{document}
\definecolor{myblue}{rgb}{0.0, 0.078, 1.0}

\title{How Does Audio Influence Visual Attention in Omnidirectional Videos? Database and Model}

\author{Yuxin Zhu$^{\dag}$\thanks{This work was supported by the National Natural Science Foundation of China under Grants 62401365, 62225112, 62271312, 62132006, U24A20220. 

Yuxin Zhu, Huiyu Duan, Kaiwei Zhang, Yucheng Zhu, Xilei Zhu, Long Teng, Xiongkuo Min, and Guangtao Zhai are with the Shanghai Jiao Tong University, Shanghai 200240, China (e-mail: rye2000@sjtu.edu.cn; huiyuduan@sjtu.edu.cn; zhangkaiwei@sjtu.edu.cn; zyc420@sjtu.edu.cn;~xilei\_zhu@sjtu.edu.cn; tenglong@sjtu.edu.cn;~minxiongkuo@sjtu.edu.cn; zhaiguangtao@sjtu.edu.cn). }, Huiyu Duan$^{\dag}$\thanks{$^{\dag}$~Equal contribution.}, Kaiwei Zhang, Yucheng Zhu, Xilei Zhu, Long Teng, Xiongkuo Min$^{\ast}$,~\IEEEmembership{Member,~IEEE}, Guangtao Zhai$^{\ast}$\thanks{* Corresponding authors.},~\IEEEmembership{Fellow,~IEEE}}




\maketitle

\begin{abstract}
Understanding and predicting viewer attention in omnidirectional videos (ODVs) is crucial for enhancing user engagement in virtual and augmented reality applications. Although both audio and visual modalities are essential for saliency prediction in ODVs, the joint exploitation of these two modalities has been limited, primarily due to the absence of large-scale audio-visual saliency databases and comprehensive analyses. This paper comprehensively investigates audio-visual attention in ODVs from both subjective and objective perspectives. Specifically, we first introduce a new \underline{a}udio-\underline{v}isual \underline{s}aliency database for \underline{o}mni\underline{d}irectional \underline{v}ideos, termed AVS-ODV database, containing 162 ODVs and corresponding eye movement data collected from 60 subjects under three audio modes including mute, mono, and ambisonics. Based on the constructed AVS-ODV database, we perform an in-depth analysis of how audio influences visual attention in ODVs. To advance the research on audio-visual saliency prediction for ODVs, we further establish a new benchmark based on the AVS-ODV database by testing numerous state-of-the-art saliency models, including visual-only models and audio-visual models. In addition, given the limitations of current models, we propose an innovative \underline{omni}directional \underline{a}udio-\underline{v}isual \underline{s}aliency prediction network (OmniAVS), which is built based on the U-Net architecture, and hierarchically fuses audio and visual features from the multimodal aligned embedding space. Extensive experimental results demonstrate that the proposed OmniAVS model outperforms other state-of-the-art models on both ODV AVS prediction and traditional AVS predcition tasks. The AVS-ODV database and the OmniAVS model are available at: 
\textcolor{myviolet}{\url{https://github.com/IntMeGroup/AVS-ODV}}.

\end{abstract}


\section{Introduction}
\IEEEPARstart{O}{mnidirectional} videos (ODVs) have attracted widespread attention in recent years due to their capability to provide immersive experiences by presenting omnidirectional multimodal information to users \cite{min2024perceptual,duan2018perceptual, zhu2023perceptual}. 
This necessitates efficient processing techniques to keep and improve the immersive and interactive experience of ODVs, which further stress the importance of understanding and predicting viewers' attention in ODVs, \textit{i.e.}, saliency prediction \cite{8269807,8960364}.
Efficient and effective saliency prediction algorithms can help optimize content delivery for streaming applications by prioritizing high-quality streaming in areas of high visual interest to enhance transmission efficiency, and can facilitate efficient rendering of high-resolution textures in salient regions to reduce processing demands. Furthermore, saliency prediction can also assist user interaction and improve user quality of experience (QoE) in virtual reality (VR) environments.

Saliency prediction has long been an important task in image processing and computer vision \cite{drostejiao2020,  sun2023influence,duan2019visual,9770033}.
Many saliency databases for traditional flat images and videos have been established \cite{5459462,aclnet,10.1109/TPAMI.2014.2366154,8451338,10.1145/3304109.3325818}, and numerous corresponding image and video saliency models have been proposed \cite{730558,10.5555/2976456.2976525, SR,SMVJ,SUN,PQFT,Judd,SWD,Murray,HFT,sun2024visual}, which mainly focus on the pure visual modality. 
With the advancement of multimodal research, some recent studies have also explored the impact of audio on visual attention \cite{Mital2011ClusteringOG,Coutrot2014HowSF,Coutrot2016,7457921,Gygli2014CreatingSF,10.1016/j.image.2015.08.004} and have proposed corresponding audio-visual saliency models for traditional flat videos \cite{dave,tsiami2020stavis}. 

\begin{figure}[t!]
\centering
\begin{minipage}{1\linewidth}
		\centering
		\includegraphics[width=0.29\linewidth]{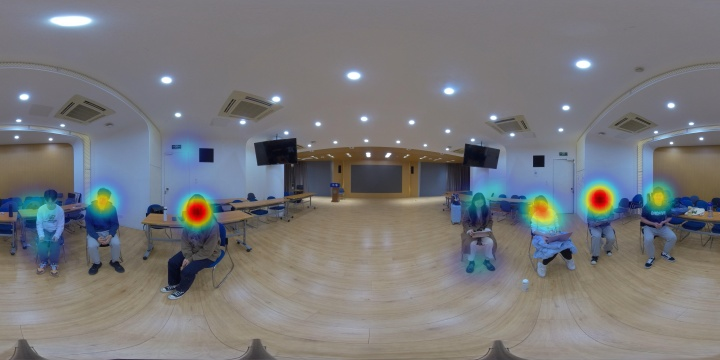}\hspace{-0.02cm}
  \includegraphics[width=0.29\linewidth]{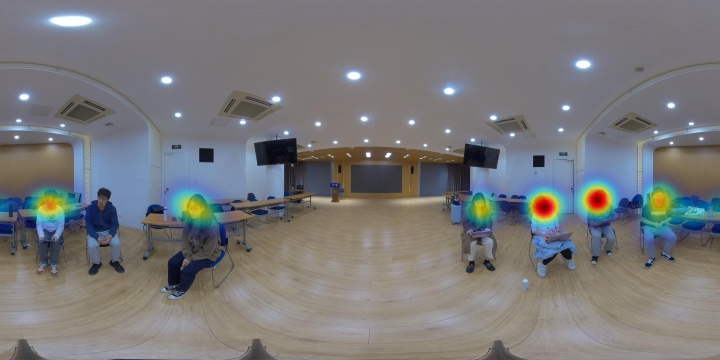}\hspace{-0.02cm}
  \includegraphics[width=0.29\linewidth]{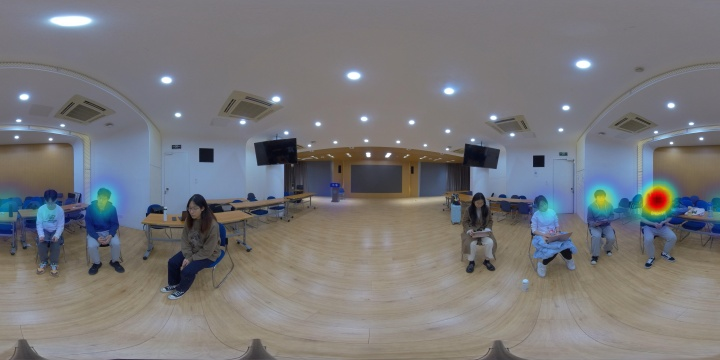}
\end{minipage}\\ \vspace{-0.28cm}
\subfloat[mute]{\label{fig:gt-mute}\includegraphics[width=0.29\linewidth]{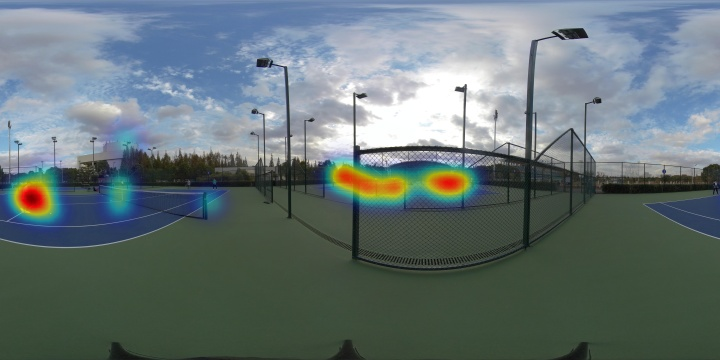}} \hspace{-0.02cm}
\subfloat[mono]{\label{fig:gt-mono}\includegraphics[width=0.29\linewidth]{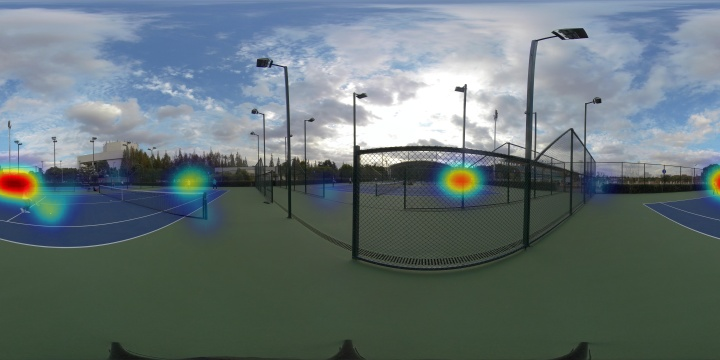}}\hspace{-0.02cm}
\subfloat[ambisonics]{\label{fig:gt-ambi}\includegraphics[width=0.29\linewidth]{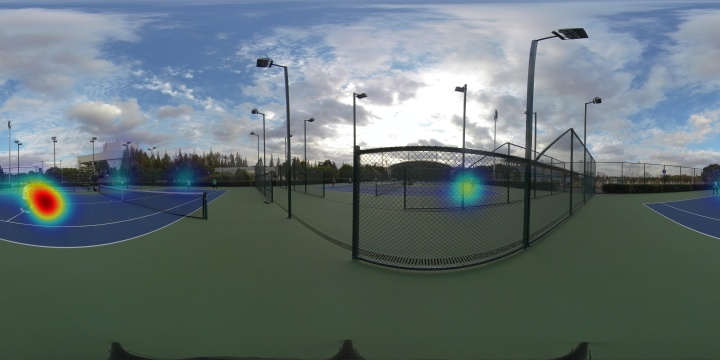}} \vspace{-0.15cm}
\caption{The heatmaps of user attention on video frames of two videos in AVS-ODV database under different audio modes.}
\label{fig:intro-gt}
\vspace{-0.3cm}
\end{figure}

Saliency prediction for ODVs is more challenging than for traditional flat videos \cite{duan2024quick}, due to the wider perspectives and richer scenes. Therefore, many recent studies have established saliency databases and designed saliency models for ODVs, which mainly focus on the pure visual modality \cite{8418756,Zhang_2018_ECCV}. However, user visual attention in ODVs is significantly influenced by audio, especially ambisonics (a full-surround audio technology widely used in ODVs). As illustrated in Fig. \ref{fig:intro-gt}, human visual attention within the same scene varies significantly under the mute, mono, and ambisonics audio modes, and viewers demonstrate increased sensitivity to sound sources under the ambisonics mode. However, the audio-visual attention and saliency prediction for ODVs have rarely been studied in the current literature.


Three existing ODV saliency databases have incorporated audio information during their construction \cite{9897737,10224292,9105956}, however, these databases have limitations such as small dataset size, lack of research on the impact of spatial audio, or only containing head movement data. Moreover, the prediction models derived from these databases, including AVS360 \cite{avs360} and SVGC-AVA \cite{10224292}, are limited by relying solely on audio energy maps without incorporating the semantic information of the audio. Therefore, there is an urgent need to develop large-scale ODV eye-tracking databases with spatial audio, and design corresponding saliency models.

In this work, we focus on investigating and modeling users' visual attention in omnidirectional videos with audio. Our contributions are fourfold:

\subsubsection{Constructing the AVS-ODV Database} 
We build a database for audio-visual saliency prediction in ODVs. This database encompasses 162 videos captured in diverse scenes. To better understand how auditory stimuli influence viewer attention, we categorize the videos into three types based on the number of visually salient objects and sound sources present in the videos. A total of 60 participants are recruited to participate in the experiment, which are randomly divided into three groups. Eye and head movement data from the three groups are collected and processed to derive fixation maps and saliency maps under three audio settings, including mute, mono, and ambisonics, respectively. To the best of our knowledge, AVS-ODV is the largest ODV audio-visual saliency prediction database, which is expected to better help attention modeling for ODVs.

\subsubsection{Comprehensive Analysis of Attention in Audio-Visual ODVs}
Based on the established AVS-ODV database, we comprehensively conduct
 qualitative and quantitative analysis of the attention mechanism in the omnidirectional audio-visual context. Our findings show that (1) the presence of sound, especially spatial audio, significantly influences visual attention; (2) in scenarios with multiple salient objects but only one sound source, the impact of audio on visual attention is particularly pronounced. This in-depth analysis provides valuable insights for subsequent attention modeling efforts.

\subsubsection{New Benchmarks for Audio-Visual Saliency Prediction in ODVs} We benchmark a total of 21 models, including 10 classical models based on handcrafted features and 11 state-of-the-art deep learning models, on the AVS-ODV database and analyze their performance. Given the absence of existing algorithms for the LJ-ODV database \cite{9897737}, we also establish a benchmark for it. We believe these benchmarks will serve as valuable resources to promote further research on audio-visual saliency prediction for ODVs.

\subsubsection{A Novel Audio-Visual Saliency Prediction Model for ODVs} We develop a new omnidirectional audio-visual saliency prediction network based on the U-Net architecture, termed OmniAVS net. To effectively integrate audio-visual features, we utilize the joint embedding space provided by the multimodal foundation model, ImageBind \cite{girdhar2023imagebind}. Specifically, we leverage ImageBind's audio and visual feature encoders to map visual and auditory features into an aligned embedding space. Building based on the U-Net architecture, we leverage multi-level visual features, and progressively inject auditory features into visual features via multi-stage cross attentions during the decoding process. Experimental results demonstrate that our model performs the best on the AVS-ODV database, and two other omnidirectional audio-visual saliency databases, \textit{i.e.}, YQ-ODV \cite{10224292} and LJ-ODV \cite{9897737}. Additionally, the visual component of OmniAVS also excels on two visual-only saliency databases for ODVs, \textit{i.e.}, PVS-HM \cite{8418756} and Zhang-ODV \cite{Zhang_2018_ECCV}. Moreover, our OmniAVS net also achieves the best performance on a combined database comprising six traditional flat audio-visual saliency databases \cite{Mital2011ClusteringOG,Coutrot2014HowSF,Coutrot2016,7457921,Gygli2014CreatingSF,10.1016/j.image.2015.08.004}, which further demonstrates the generalization ability and superiority of our proposed model.

This work significantly extends our ICIG paper \cite{zhu2023audio} in several key aspects. Firstly, we provide additional details on the construction of the AVS-ODV database and analyze the diversity of its content. Secondly, we offer a more comprehensive qualitative and quantitative analysis of visual attention in ODVs based on the AVS-ODV database. Thirdly, we establish benchmark libraries for each audio mode in AVS-ODV, incorporating a wider range of saliency models. Fourthly, we introduce a new audio-visual saliency prediction algorithm, OmniAVS, and conduct extensive experiments and analyses to evaluate its performance. 

The rest of the paper is organized as follows. Section \ref{related} provides a literature review of saliency databases and models for ODVs. Section \ref{dataset} details the construction and analysis of the proposed AVS-ODV database. Section \ref{model} describes our proposed OmniAVS model. Section \ref{experiments} offers experimental studies of OmniAVS. Section \ref{conclusion} concludes the whole paper.

\section{Related Work}\label{related}
\subsection{Saliency Databases for ODVs}
Most existing ODV saliency databases only conduct eye-tracking experiments on omnidirectional videos without audio tracks, including PVS-HM \cite{8418756} and Zhang-ODV \cite{Zhang_2018_ECCV}, \textit{etc}. PVS-HM includes the head and eye movement data from 58 participants in 76 ODVs, while Zhang-ODV comprises 104 videos with eye movement data from 27 participants. 

The ODV saliency databases that include audio information remain limited. Li \textit{et al.} \cite{9897737} developed a database that includes 43 ODVs with stereo audio information, along with head and eye movement data from 30 participants. 360AV-HM \cite{8551543} is the first database that incorporates ambisonics. However, it only contains head movement data and includes just 12 videos, which limits its scope for comprehensive research on audio-visual saliency algorithms. Subsequently, Yang \textit{et al.} \cite{10224292} introduced a database that also considers ambisonics, which includes a slightly larger collection of 57 ODVs, along with head and eye movement data. However, its size remains relatively modest, which is insufficient for comprehensive research needs.

\subsection{Saliency Prediction Models for ODVs}
Although many ODV saliency models have been developed, they primarily focus solely on visual information. Cheng \textit{et al.} \cite{8578252} introduced a weakly-supervised spatial-temporal network for saliency prediction in ODVs without audio. It features the innovative Cube Padding technique, which effectively addresses the distortion issues inherent in ODV processing. Zhang \textit{et al.} \cite{Zhang_2018_ECCV} employed a spherical U-Net architecture specifically tailored for ODVs. Xu \textit{et al.} \cite{8418756} introduced a deep reinforcement learning approach for predicting head movement data in ODVs. Dahou \textit{et al.} \cite{10.1007/978-3-030-68796-0_22} proposed ATSal model, which combines global visual attention with local patch information using an attention mechanism to enhance saliency prediction in ODVs.
These existing visual-only algorithms have made significant progress in modeling viewer attention in ODVs. However, their limitation is that they rely exclusively on visual information, which overlooks the rich auditory context that often accompanies ODVs.

AVS360 \cite{avs360} is the first audio-visual saliency prediction model for ODVs. It utilizes Cube Padding within its 3D ResNet architecture to mitigate geometric distortions in visual feature and fuses audio and visual modalities via simple feature concatenation. Despite its innovative approach, the simple concatenation strategy without prior alignment between audio and visual features may result in suboptimal fusion problem. SVGC-AVA \cite{10224292} is distinguished by its innovative use of Spherical Vector-Based Graph Convolution (SVGC), which ensures distortion-free visual information processing. The model utilizes AEM to represent the audio modality, and integrates the attention mechanism \cite{10.5555/3295222.3295349} as a part of its audio-visual attention strategy to explore both self-modal and cross-modal correlations. The SVGC-AVA model has advanced audio-visual saliency prediction performance in ODVs but is still limited by its neglect of video temporal information and audio semantic features.

\section{AVS-ODV Database}\label{dataset}
To comprehensively study the effect of sound on visual attention in ODVs, we construct an audio-visual saliency database for ODVs, termed AVS-ODV. We first collect 162 ODV clips with spatial audio and then perform eye-tracking experiments to obtain eye movement data under three audio modes, including mute, mono, and ambisonics, respectively.

\subsection{Data Collection}
\subsubsection{Stimuli}
\begin{figure}[t!]
\setlength{\abovecaptionskip}{0cm} 
\setlength{\belowcaptionskip}{-0.1cm}
\centering
\includegraphics[width=0.75\linewidth]{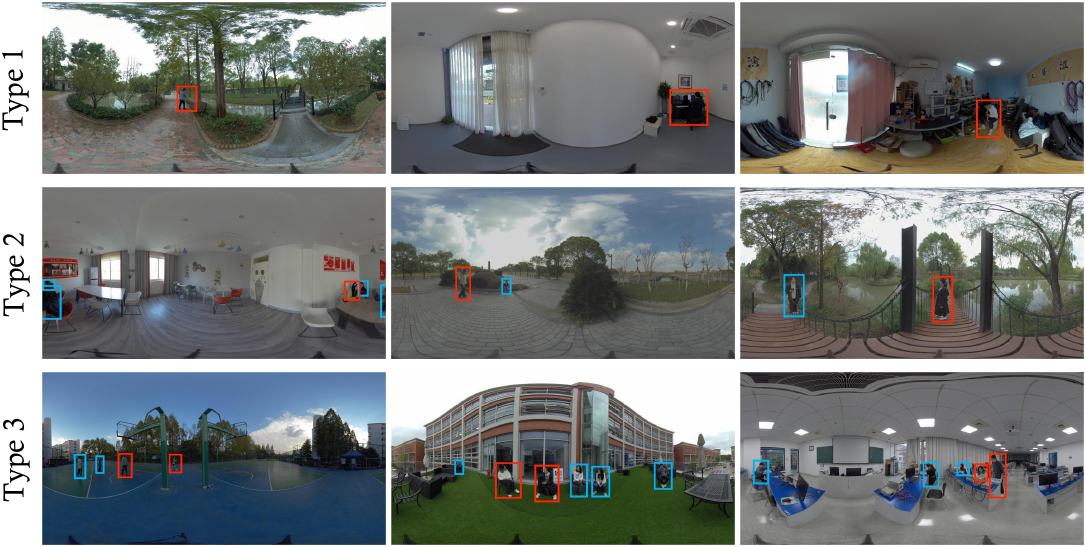}
\caption{Sample frames of different types of videos in AVS-ODV. Red boxes represent sound sources, and blue boxes represent the other salient objects.}
\label{fig:stimuli-sample}
\vspace{-0.2cm}
\end{figure}

We use a professional VR camera Insta360 Pro2 \cite{insta360} to capture 162 ODV clips in various scenes. Each ODV features a resolution of 7680$\times$3840 in the ERP format, a frame rate of 29.97 fps, and a standardized duration of 15 seconds. All videos include first-order ambisonics in B-format with an audio sampling rate of 48,000Hz. The four channels of the first-order ambisonics are spherical audio components: ``W" captures omnidirectional sound pressure, while ``X", ``Y" and ``Z" encode three-dimensional directional sound information along the front-back, left-right and up-down axes, respectively. Since the audio in our videos is entirely captured during filming, it is inherently aligned with the visual content. The audio is either directly associated with visually salient objects or comes from non-salient sources that are not immediately visible, such as birds calling from the background trees. This alignment between audio and visual stimuli is crucial, as mismatched audio-visual stimuli have been shown to diminish the effectiveness of visual attention~\cite{marighetto2017audio}. The objects of interest in our database encompass both top-down and bottom-up stimuli. Top-down stimuli include semantically significant entities such as individuals engaged in specific activities (e.g., a speaker delivering a lecture) or contextually relevant objects. These stimuli attract attention based on their relevance or context. Bottom-up stimuli, on the other hand, refer to sensory-driven inputs that capture attention due to their inherent properties, such as the ambient noise of a crowded public event, which may not have a clear visual source but can still influence attention.
We hypothesize that visual attention is associated with salient objects and sound sources, where salient objects mainly refer to people or moving objects. Therefore, we categorize our videos into three types based on the number of salient objects and sound sources, \textit{i.e.}, videos with a single salient object where the sound source is that object (type 1), videos with multiple salient objects but only one of them is the sound source (type 2), and videos with multiple salient objects and multiple sound sources (type 3). Fig. \ref{fig:stimuli-sample} displays representative video frames from these three types of videos. The red boxes indicate sound sources, while the blue boxes indicate the other salient objects.

\begin{figure*}[t!]
\setlength{\abovecaptionskip}{0cm}
\setlength{\belowcaptionskip}{-0.2cm}
\centering
\includegraphics[width=0.88\linewidth]{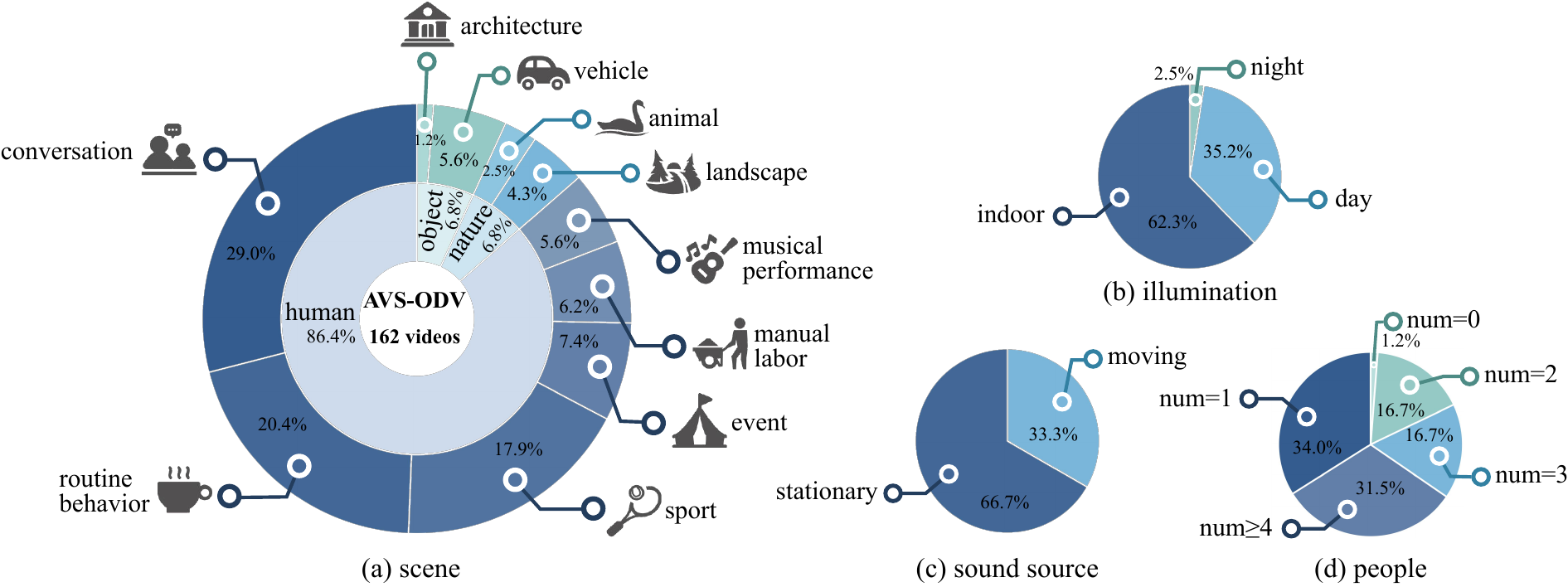}\vspace{-0.2cm}
\caption{Statistical overview of key features in the AVS-ODV database.\label{fig:stimuli-statistics}}
\vspace{-0.05cm}
\end{figure*}
\begin{figure*}[t!]
\vspace{-0.2cm}
\centering
\subfloat[Brightness]{\label{fig:Brightness}\includegraphics[height=2cm]{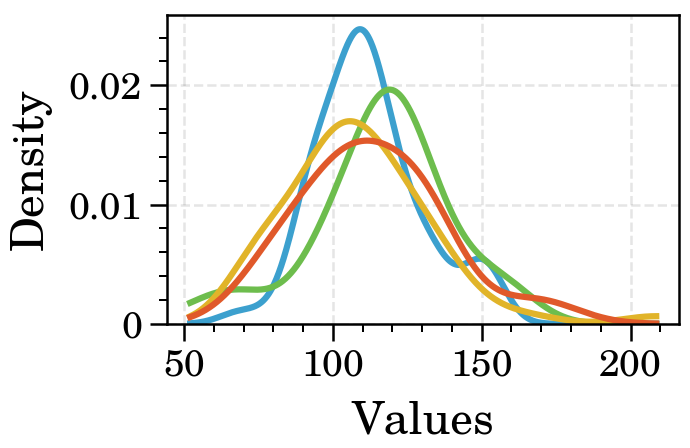}} \hspace{0.06cm}
\subfloat[Contrast]{\label{fig:Contrast}\includegraphics[height=2cm]{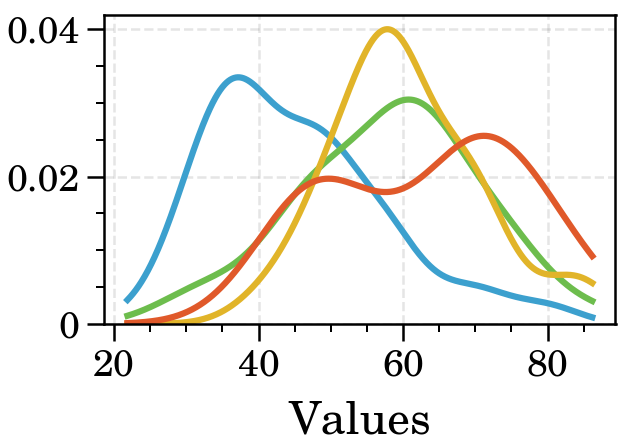}}\hspace{0.06cm}
\subfloat[SI]{\label{fig:SI}\includegraphics[height=2cm]{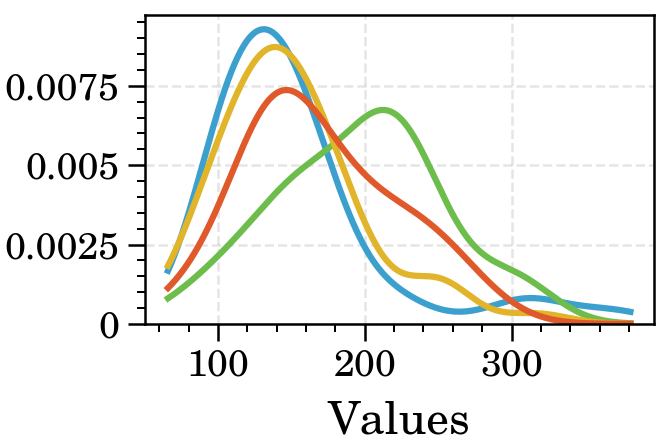}} \hspace{0.06cm}
\subfloat[SEF]{\label{fig:SEF}\includegraphics[height=2cm]{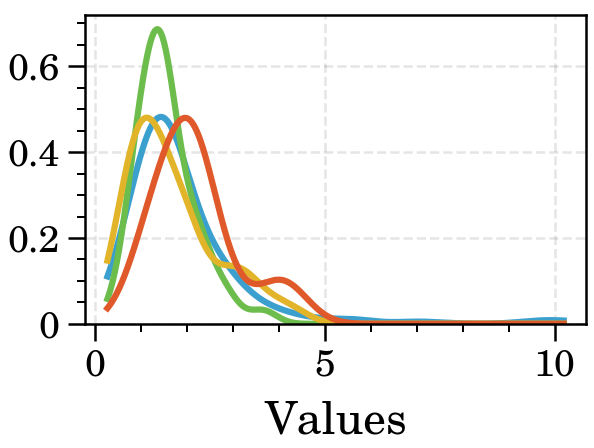}}\hspace{0.06cm} 
\subfloat[ZCR]{\label{fig:ZCR}\includegraphics[height=2cm]{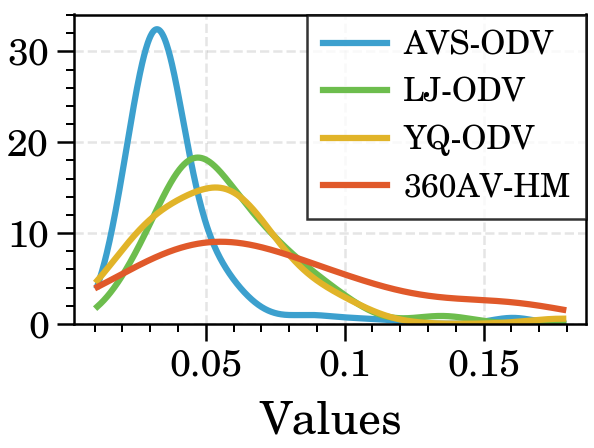}}\\\vspace{-2.5pt}
\caption{Comparison of video and audio attributes of four databases: AVS-ODV, LJ-ODV \cite{9897737}, YQ-ODV \cite{10224292}, 360AV-HM \cite{9105956}.}
\label{fig:stimuli-analysis}
\vspace{-0.3cm}
\end{figure*}

To demonstrate the diversity of our database, we provide a statistical overview of the different categories of scenes within the AVS-ODV database. As shown in Fig.\ref{fig:stimuli-statistics}, the categories are divided into human activities (e.g., conversations, sports, routine behaviors), natural scenes (e.g., animals, landscapes), and object-related scenes (e.g., vehicles, architectures). This distribution highlights the wide variety of real-world contexts covered by the database. Additionally, we analyze the distribution of lighting conditions, the movement of sound sources, and the number of people in each video, as shown in Fig. \ref{fig:stimuli-statistics}. This analysis highlights the database's extensive diversity, capturing variations in environmental, temporal, and social factors that reflect the complexities of real-world scenarios. We also compare the AVS-ODV database with three other ODV audio-visual saliency databases, including the 360AV-HM database \cite{9105956}, the LJ-ODV database \cite{9897737}, and the YQ-ODV database \cite{10224292}. Table \ref{tab:basicinfo} summarizes the main features of these four databases, highlighting the advantages of AVS-ODV. 
We further compute visual attributes, including brightness, contrast and spatial perceptual information (SI) \cite{siti}, as well as audio attributes, including short-term energy fluctuation (SEF) and zero-crossing rate (ZCR) \cite{giannakopoulos2014introduction}. Fig. \ref{fig:stimuli-analysis} illustrates that AVS-ODV has a wide range of distributions across attributes, similar to the other databases.

\begin{table}[t!]
\setlength{\abovecaptionskip}{0cm} 
\setlength{\belowcaptionskip}{-0.4cm}
\caption{Basic Information of Four Databases: 360AV-HM, LJ-ODV, YQ-ODV, AVS-ODV. ``HM" and ``EM" represent head movement and eye movement respectively.\label{tab:basicinfo}}
\setlength\tabcolsep{4pt} 
\centering
\begin{adjustbox}{width=0.48\textwidth}
\begin{tabular}{@{}l|lllll@{}}
\toprule
Database & Num & Resolution& Duration & Audio  & Data Type \\ \midrule
360AV-HM\cite{9105956} & 15 & 3840$\times$1920 & 25s & mute, mono, ambisonics & HM\\
LJ-ODV\cite{9897737} & 43 & 3840$\times$1920 & 15s & mute, stereo &HM+EM \\
YQ-ODV\cite{10224292} & 57 & 3840$\times$1920 & 28s & mute, mono, ambisonics &HM+EM\\ 
AVS-ODV (ours) & 162 & 7680$\times$3840& 15s & mute, mono, ambisonics  & HM+EM \\
\bottomrule
\end{tabular}
\end{adjustbox}
\vspace{-0.3cm}
\end{table}

\subsubsection{Apparatus}
We utilize a HTC VIVE Pro Eye \cite{VIVE,duan2022confusing} to play ODVs and record head and eye movement data. It is a high-performance VR Head-Mounted Display (VR-HMD) equipped with an eye tracker that employs Tobii eye-tracking technology. The eye-tracking frequency is 120Hz, allowing for precise tracking of the viewer's head and eye movements. We develop a software using Unity \cite{Unity}, which is supplemented by the Media Player plugin from AVPro Video \cite{avpro} and Facebook Audio 360 \cite{facebookaudio}. 

\subsubsection{Subjects}
We recruit 60 participants for our study, including 26 males and 34 females, and all subjects have normal or corrected-to-normal vision. We divide all subjects into three groups, ensuring a balanced gender ratio in each group, and collect data from 20 subjects for each audio mode (mute, mono, and ambisonics).

\subsubsection{Procedure}
During the experiment, participants are seated in a swivel chair. They are first instructed to wear the VR-HMD and watch an omnidirectional video sample to adapt to the virtual environment. Subsequently, eye-tracking calibration is performed to ensure the accuracy of data tracking. The participants are then encouraged to explore the presented videos freely. To ensure consistency across viewers, we fix the initial position of the VR-HMD's viewport center at the 180$^\circ$ longitude of the ODV. The video sequence is shuffled, and each participant views each video only once. To avoid visual fatigue, we divide the 162 videos into two parts with a break in the middle. There is also a two-second black-screen interval between each two videos. The experiment lasts approximately 50 to 60 minutes per participant.

\subsection{Data Processing and Analysis}\label{subsec:sim}
\subsubsection{Data Processing}
The raw eye movement data is collected in the three-dimensional Cartesian coordinate $(x, y, z)$, where fixation points scatter on the surface of a sphere with radius $r = 1$. The raw head movement data are recorded in terms of $(\theta_{pitch}, \theta_{yaw}, \theta_{roll})$. Both types of raw data are first transformed into orientation vectors in latitude and longitude coordinate format. Fixations are extracted by first filtering out saccades, which are fast movements that do not indicate user attention. Velocity and acceleration are derived from the orientation vectors, and saccades are identified and removed if their velocity exceeds 20$^\circ$/s or their acceleration exceeds 50$^\circ$/s$^2$~\cite{10.1145/3304109.3325820}. Then, the orientation vectors are projected into pixel coordinates. To remove random or isolated fixations, the DBSCAN algorithm~\cite{10.5555/3001460.3001507} is applied. DBSCAN clusters high-density fixation points while filtering out noise, preserving core fixations even in less dense areas. The fixation maps are generated by aggregating fixation points from all users on a frame-by-frame basis. Finally, a Gaussian filter with a visual angle of 3.34$^\circ$ is applied to the fixation maps on the sphere to generate saliency maps~\cite{Gutirrez2018ToolboxAD}, creating smooth and continuous regions of visual attention.

\subsubsection{Qualitative Data Analysis}
\begin{figure}[t!]
\vspace{-0.25cm}
\setlength{\abovecaptionskip}{0cm}
\setlength{\belowcaptionskip}{-0.2cm}
\centering
\subfloat[mute]{\label{fig:global_mute}\includegraphics[height=2.86cm]{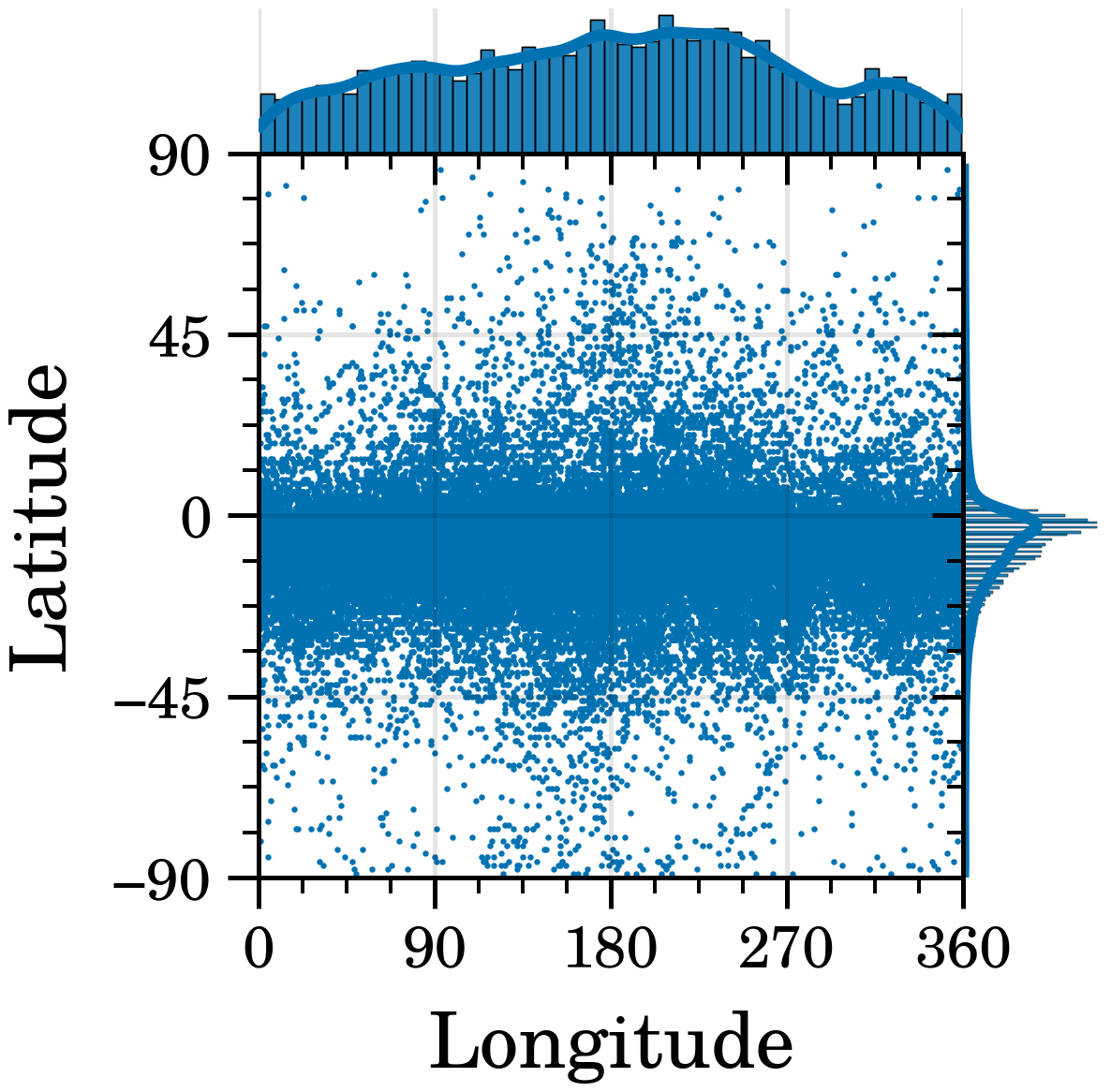}} 
\subfloat[mono]{\label{fig:global_mono}\includegraphics[height=2.86cm]{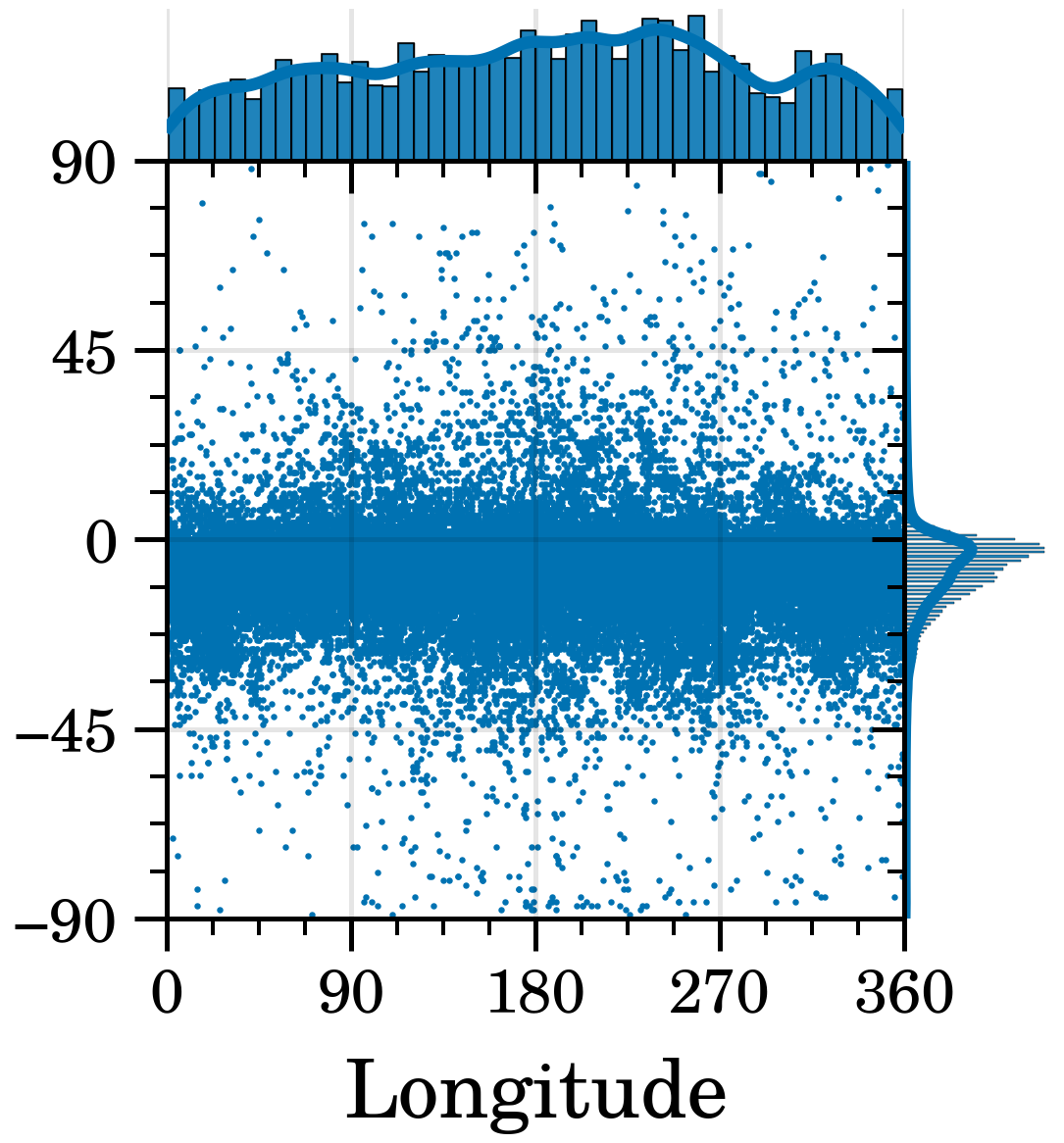}}
\subfloat[ambisonics]{\label{fig:global_ambi}\includegraphics[height=2.86cm]{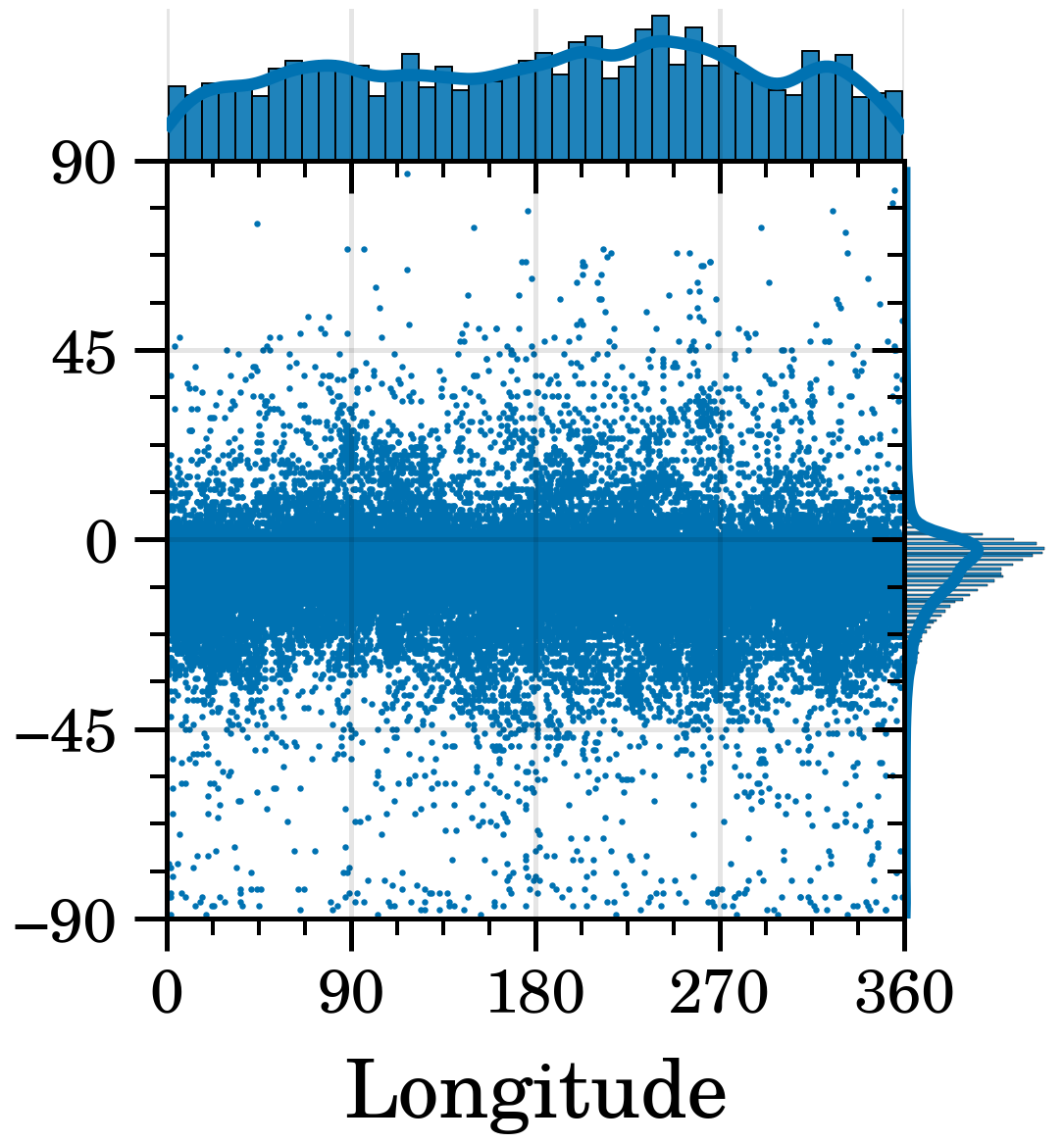}} 
\caption{Distribution of fixations along the latitude and longitude axes under three audio modes.}
\label{fig:global-joint}
\vspace{-0.4cm}
\end{figure}

First, we analyze the global differences in attention across three audio modes. Fixations from all video frames in AVS-ODV are overlaid for each mode, resulting in the joint distribution maps shown in Fig. \ref{fig:global-joint}. Across all modes, the distribution of visual attention along the latitude exhibits an ``Equator Bias". For longitude, the observer's attention is most concentrated near the initial position (at the 180$^\circ$ meridian). The mute mode shows the highest number of fixation points near the initial observation area. In contrast, the ambisonics mode displays a broader distribution of attention along longitude, with numerous fixation points even far from the initial position. This suggests that sound, particularly ambisonics, can direct the observer's attention to a broader area than just near the initial viewing area.

\begin{figure}[t!]
\vspace{-0.25cm}
\setlength{\abovecaptionskip}{0.1cm}
\setlength{\belowcaptionskip}{0cm}
\centering
\subfloat[Type 1]{\label{fig:fixaem1}\includegraphics[height=0.25\linewidth]{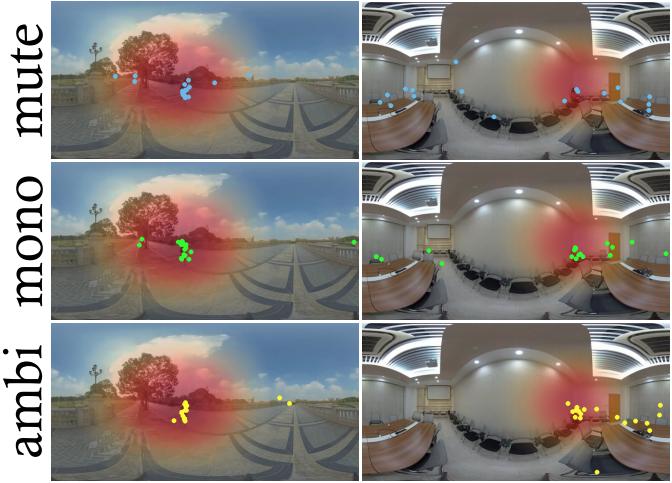}} 
\subfloat[Type 2]{\label{fig:fixaem2}\includegraphics[height=0.25\linewidth]{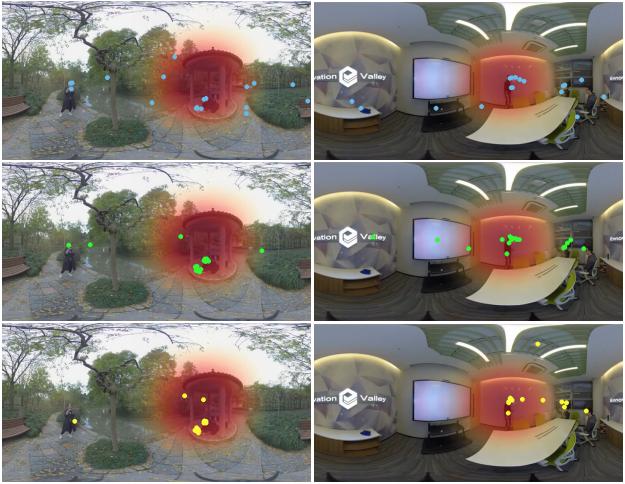}}
\subfloat[Type 3]{\label{fig:fixaem3}\includegraphics[height=0.25\linewidth]{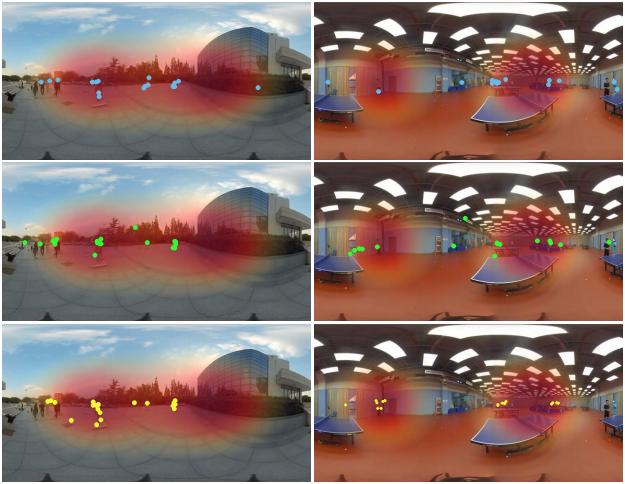}} 
\caption{Fixations under three audio modes for three types of ODVs. The audio energy distribution is depicted in red. The {\color[HTML]{69B9E8}blue}, {\color[HTML]{32FA32}green}, and {\color[HTML]{F2EA08} yellow} dots indicate the fixations under mute, mono, and ambisonics modes, respectively.}
\label{fig:fix-aem}
\vspace{-0.36cm}
\end{figure}

\begin{figure*}[t!]
\setlength{\abovecaptionskip}{0.1cm}
\setlength{\belowcaptionskip}{0cm}
\vspace{-0.3cm}
\centering
\subfloat[Type 1]{\label{fig:1_08_polar}\includegraphics[width=0.325\linewidth]{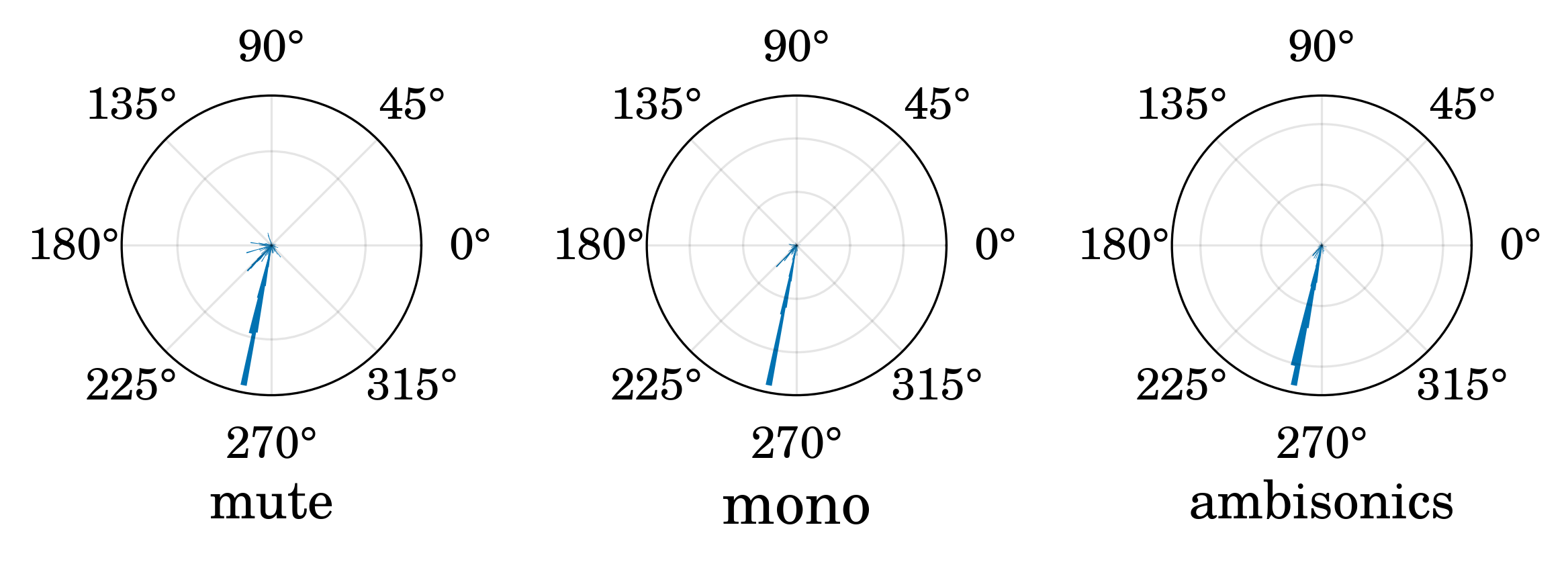}} \hspace{1pt}
\subfloat[Type 2]{\label{fig:2_26_I_S_polar}\includegraphics[width=0.325\linewidth]{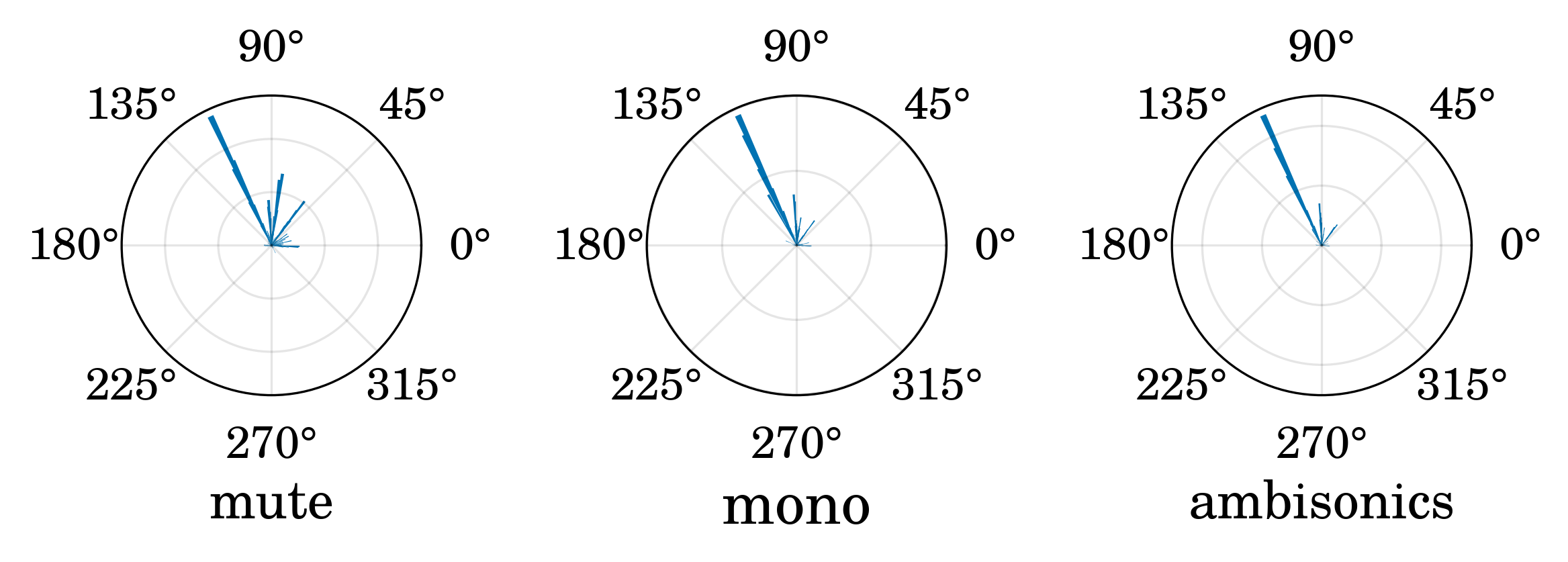}}\hspace{1pt}
\subfloat[Type 3]{\label{fig:3_47_I_S_polar}\includegraphics[width=0.325\linewidth]{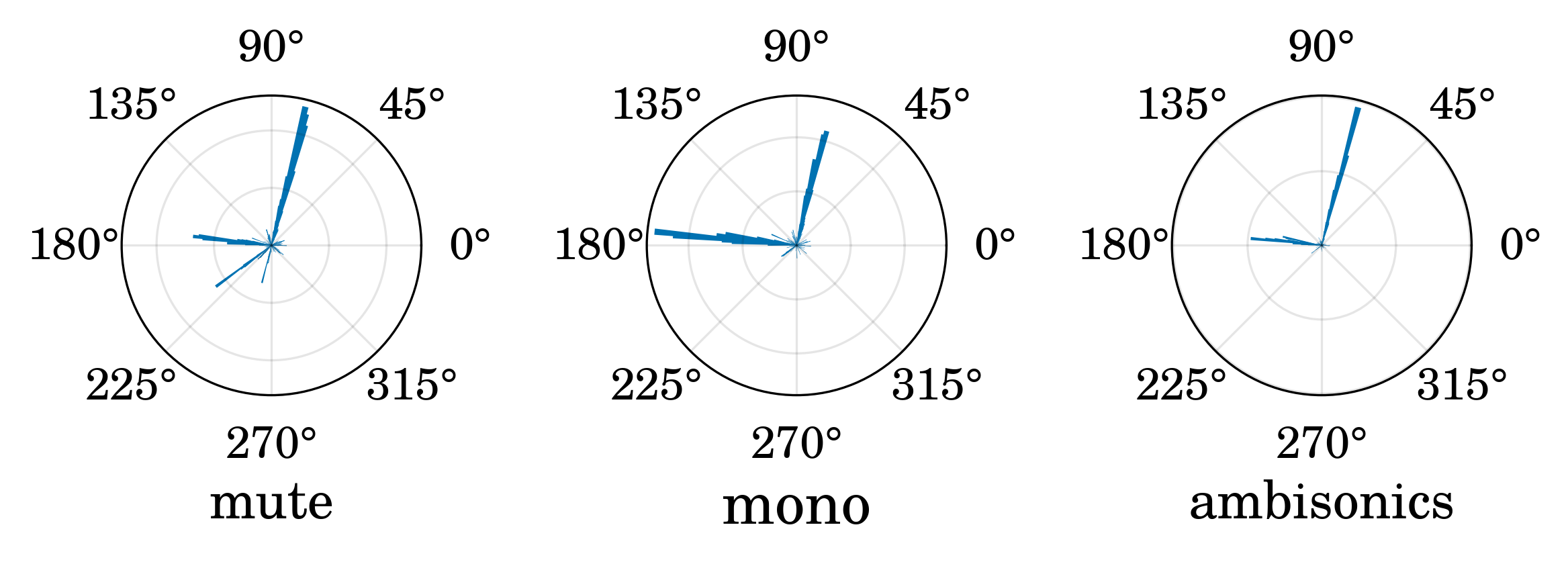}} 
\caption{Distribution of fixations along the longitude across three audio modes for three types of ODVs.}
\label{fig:polar}
\vspace{-0.2cm}
\end{figure*}

We further investigate the effect of audio on visual attention in ODVs for different video types. Fig. \ref{fig:fix-aem} displays one frame from two videos for each type, annotated with the corresponding fixations under the three modes. The audio energy distribution \cite{morgadoNIPS18} is depicted in red to visualize the sound sources. For type 1 videos, fixation distributions across all audio modes converge mainly on the video's salient object (also the sound source). The highest concentration of fixations is observed under the ambisonics mode, followed by the mono mode. In type 2 videos, the majority of fixations under the ambisonics mode are concentrated on the sound source, whereas under the mono mode, a small proportion of fixations are distributed to other salient objects. Under the mute mode, attention is distributed among various salient objects or other elements. In type 3 videos, attention is distributed among sound sources under the ambisonics mode, while it is more scattered under the other two audio modes. We also select one video for each type and plot the distribution of fixations along longitude in Fig. \ref{fig:polar}, which confirms our finding.

\subsubsection{Quantitative Data Analysis}
We adopt three most commonly used metrics, including normalized scanpath saliency (NSS) \cite{PETERS20052397}, similarity (SIM), and linear correlation coefficient (CC), to evaluate the consistency of visual attention across different audio modes based on saliency maps and fixation maps. To correct for geometric distortions at the poles in the ERP format, we apply a weighting factor of $\sin{(90^\circ-\alpha)}$ for $\alpha \in [-90^\circ,90^\circ]$ to the saliency maps, where $\alpha$ represents the latitude angle. Inspired by the work of Song \textit{et al.} \cite{Song}, we adopt a randomized approach to create a baseline for statistical comparison. We randomly divide all 60 subjects into two groups and calculate the consistency of attention between the two groups. This procedure is repeated five times, and the average is taken as the baseline. We further calculate the consistency of visual attention between each pair of audio modes. If the results are lower than the baseline, it indicates that the audio modes have a significant influence on visual attention. Moreover, the lower correlation value represents a larger difference of the influence on visual attention. According to the results shown in Table \ref{tab:consistency}, the consistency of attention between the mute and ambisonics modes is the lowest, followed by that between the mute and mono modes. This indicates that the presence of sound, particularly spatial audio, significantly influences visual attention.
Conversely, the consistency of attention between the mono and ambisonics modes is relatively high, suggesting that the orientation of audio has less influence on visual attention than the presence of audio. Furthermore, the visual attention consistency for type 2 ODVs is the lowest across different audio modes. This suggests that audio has the most pronounced effect on visual attention in videos with multiple salient objects but only one sound source.

\begin{table}[t!]
\setlength{\abovecaptionskip}{0cm}
\setlength{\belowcaptionskip}{-0.2cm}
\caption{Consistency scores for three types of ODVs. The values in parentheses represent the percentage change in consistency between the two audio modes relative to the baseline.\label{tab:consistency}}
\setlength\tabcolsep{4pt} 
\centering
\begin{adjustbox}{width=0.44\textwidth}
\begin{tabular}{@{}clllll@{}}
\toprule
\multicolumn{1}{c|}{Metrics} &  \multicolumn{1}{c|}{Type} & \multicolumn{1}{l|}{Baseline}& \multicolumn{1}{l}{Mute-Mono} & \multicolumn{1}{l}{Mute-Ambisonics} & \multicolumn{1}{l}{Mono-Ambisonics} \\ \midrule
\multicolumn{1}{c|}{} & \multicolumn{1}{c|}{1} & \multicolumn{1}{l|}{6.4532} & {\color[HTML]{000000} 5.7831 (-10.38\%)} & 5.8007 (-10.11\%) & 6.9125 (+7.12\%) \\
\multicolumn{1}{c|}{} & \multicolumn{1}{c|}{2} & \multicolumn{1}{l|}{6.2414} & 5.5409 (-11.22\%) & 5.4335 (-12.94\%) & 6.6417 (+6.41\%) \\
\multicolumn{1}{c|}{} & \multicolumn{1}{c|}{3} & \multicolumn{1}{l|}{6.2032} & 5.6313 (-9.22\%) & 5.5601 (-10.37\%) & 6.4644 (+4.21\%) \\
\multicolumn{1}{c|}{\multirow{-4}{*}{NSS}} & \multicolumn{1}{c|}{all} & \multicolumn{1}{l|}{6.2946} & 5.6459 (-10.31\%) & 5.5892 (-11.21\%) & 6.6674 (+5.92\%) \\ \midrule
\multicolumn{1}{c|}{} & \multicolumn{1}{c|}{1} & \multicolumn{1}{l|}{0.6054} & 0.5200 (-14.11\%) & 0.5085 (-16.01\%) & 0.5774 (-4.63\%) \\
\multicolumn{1}{c|}{} & \multicolumn{1}{c|}{2} & \multicolumn{1}{l|}{0.6181} & 0.5286 (-14.48\%) & 0.5030 (-18.62\%) & 0.5835 (-5.60\%) \\
\multicolumn{1}{c|}{} & \multicolumn{1}{c|}{3} & \multicolumn{1}{l|}{0.6292} & 0.5502 (-12.56\%) & 0.5312 (-15.58\%) & 0.5919 (-5.93\%) \\
\multicolumn{1}{c|}{\multirow{-4}{*}{SIM}} & \multicolumn{1}{c|}{all} & \multicolumn{1}{l|}{0.6178} & 0.5331 (-13.71\%) & 0.5140 (-16.80\%) & 0.5844 (-5.41\%) \\ \midrule
\multicolumn{1}{c|}{} & \multicolumn{1}{c|}{1} & \multicolumn{1}{l|}{0.8336} & 0.7392 (-11.32\%) & 0.7252 (-13.00\%) & 0.7971 (-4.38\%) \\
\multicolumn{1}{c|}{} & \multicolumn{1}{c|}{2} & \multicolumn{1}{l|}{0.8284} & 0.7214 (-12.92\%) & 0.6918 (-16.49\%) & 0.7876 (-4.93\%) \\
\multicolumn{1}{c|}{} & \multicolumn{1}{c|}{3} & \multicolumn{1}{l|}{0.8275} & 0.7354 (-11.13\%) & 0.7120 (-13.96\%) & 0.7805 (-5.68\%) \\
\multicolumn{1}{c|}{\multirow{-4}{*}{CC}} & \multicolumn{1}{c|}{all} & \multicolumn{1}{l|}{0.8297} & 0.7315 (-11.84\%) & 0.7088 (-14.57\%) & 0.7882 (-5.00\%) \\ \bottomrule
\end{tabular}
\end{adjustbox}
\vspace{-0.4cm}
\end{table}

\section{Proposed OmniAVS Framework}\label{model}
\begin{figure*}[t!]
\setlength{\abovecaptionskip}{0cm}
\setlength{\belowcaptionskip}{-0.2cm}
\centering
\includegraphics[width=0.96\linewidth]{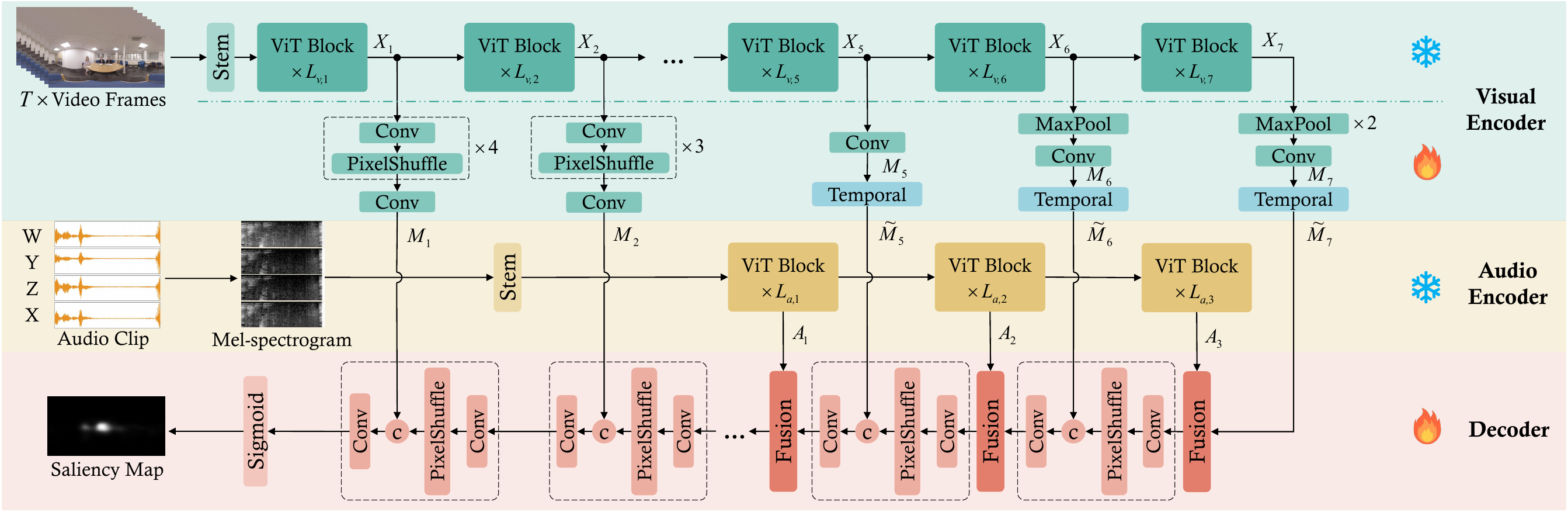}
\caption{A detailed illustration of our OmniAVS. The green part represents the visual encoder, which extracts spatial visual feature from input video frames, and conduct temporal feature aggregation. The yellow part represents the audio encoder, which extracts audio features from the input audio clip. The red part represents the decoder, which fuses the audio-visual features at multiple levels and decodes the saliency map from the multi-scale audio-visual features. }
\label{fig:AVSUNet}
\vspace{-0.2cm}
\end{figure*}

As shown in Fig. \ref{fig:AVSUNet}, we develop an omnidirectional audio-visual saliency prediction model based on U-Net\cite{10.1007/978-3-319-24574-4_28}, termed OmniAVS. It comprises a spatial visual feature representation module that extracts spatial features from input video frames, a visual feature temporal aggregation module that integrates temporal information, an audio feature representation module that extracts semantic and orientation features from the corresponding audio clip, an audio-visual feature fusion module that hierarchically infuses audio features into the visual features, and audio-visual saliency estimation blocks that decode multi-scale audio-visual features to derive saliency maps. These modules are described in detail in the following subsections. 

\vspace{-0.2cm}
\subsection{Spatial Visual Feature Representation}
We adopt ImageBind \cite{girdhar2023imagebind} as the backbone of the visual and audio feature extraction network due to its superior capability to provide an aligned audio-visual feature space. 

Historical video information within a certain time range is crucial to predict the saliency of the current video frame. Therefore, we retrospectively trace a 2-second video clip preceding the current frame, from which we uniformly sample to obtain a set of $T$ frames, including the current frame. Spatial feature extraction is performed on $T$ video frames respectively using ImageBind's ViT-H/14 vision encoder. We extract features from different stages within the 32-layer encoder blocks to encode multi-level feature representations. 
These features, denoted as $X_i$, are then up-scaled using up-sampling components constituting convolution and pixel shuffle layers, or down-scaled using max pooling layers. The final multi-level spatial feature representation at the \( i \)-th stage can be denoted as $ M_i \in \mathbb{R}^{C \times t \times h_i \times w_i} $, where \( h_i \) and \( w_i \) are the height and width of the feature map at the \( i \)th stage. For the shallow layers (\( i \in \{1, 2, 3, 4\} \)), \( t \) is set to $1$ to enhance computational efficiency, which means only the features of the current video frame are extracted for subsequent computations. For the deeper layers (\( i \in \{5,6,7\} \)), we retain the whole temporal dimension, setting \( t = T \), to facilitate the subsequent temporal aggregation module.


\vspace{-0.2cm}
\subsection{Temporal Aggregation}\label{sec:temporal}
\begin{figure}[t!]
\setlength{\belowcaptionskip}{-0.2cm}
\centering
\subfloat[3D ConvNet]{\label{fig:3D ConvNet}\includegraphics[width=0.38\linewidth]{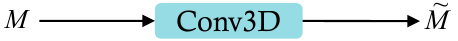}} \\\vspace{-0.3cm}
\subfloat[Spatio-Temporal GRU]{\label{fig:Spatio-Temporal GRU}\includegraphics[width=0.99\linewidth]{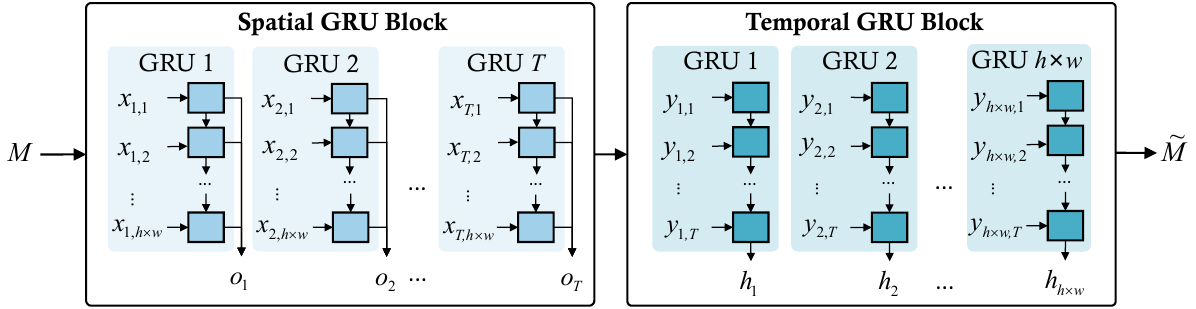}}\\\vspace{-0.3cm}
\subfloat[Spatio-Temporal Transformer]{\label{fig:Spatio-Temporal Transformer}\includegraphics[width=0.99\linewidth]{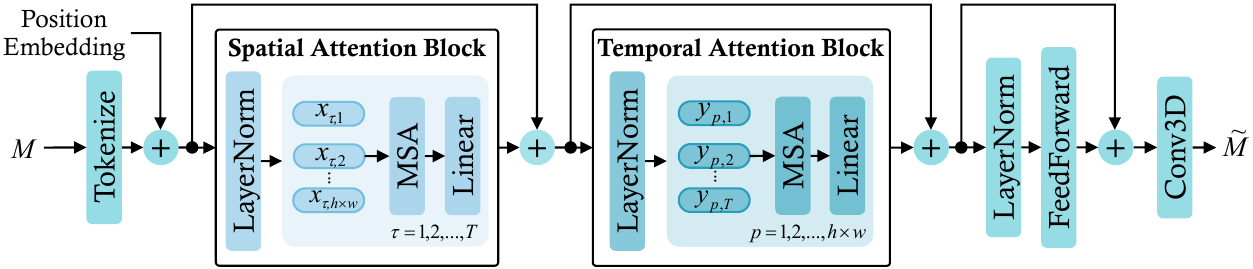}} \vspace{-0.1cm}
\caption{Three temporal aggregation methods. (a) 3D ConvNet. (b) Spatio-Temporal GRU. (c) Spatio-Temporal Transformer.}
\label{fig:temporal}
\vspace{-0.5cm}
\end{figure}
Fusing temporal information with spatial visual features is crucial for capturing the dynamic context and temporal continuity of visual scenes.
As shown in Fig. \ref{fig:temporal}, three approaches are considered for aggregating temporal features, including 3D convolutional neural network (ConvNet) \cite{6165309}, spatio-temporal gated recurrent unit (GRU), and spatio-temporal Transformer. 

\subsubsection{3D ConvNet}
3D ConvNet is an intuitive method for temporal feature fusion, which extends 2D ConvNet by adding a temporal dimension. We simply apply a 3D convNet with kernel size \( T \times 3 \times 3 \) with padding to the input visual features \( M \in \mathbb{R}^{C \times T \times h \times w} \), thereby integrating temporal information to obtain \( \tilde{M} \in \mathbb{R}^{C \times h \times w} \).

\subsubsection{Spatio-Temporal GRU}
GRU \cite{cho-etal-2014-learning} is a type of recurrent neural network well-suited for sequential data. To adapt GRUs for video feature processing, we divide this module into spatial GRU block and temporal GRU block. In the spatial GRU block, the features at the $\tau$-th time step are treated as a sequence $\{x_{\tau,p}\}_{p=1}^{h\times w}$, where  $p$ indicates the pixel index and each element in the sequence has a feature dimension of $C$. Subsequently, the outputs of the GRU cells at each time step, denoted as \(\{o_\tau\}_{\tau = 1}^T\), where $o_\tau\in \mathbb{R}^{C  \times h \times w}$, are concatenated and fed into the next stage. In the temporal GRU block, features of the $p$-th pixel in the feature map across different time steps are modeled as a sequence $\{y_{p,\tau}\}_{\tau=1}^{T}$. The final hidden states of each GRU block, denoted as \(\{h_p\}_{p=1}^{h \times w}\), are concatenated to obtain the final output of the temporal module, \(\tilde{M} \in \mathbb{R}^{C \times h \times w}\).

\subsubsection{Spatio-Temporal Transformer}
This approach is primarily inspired by ViViT \cite{9710415}. Similar to the spatio-temporal GRU, the spatio-temporal Transformer also operates in spatial and temporal stages. In the first stage, for the $\tau$-th time step, features of different spatial indexes $\{x_{\tau,p}\}_{p=1}^{h\times w}$ are treated as a sequence and processed through a Transformer encoder to compute self-attention spatially. Then, a Transformer encoder integrates features of the $p$-th spatial index across time steps, denoted as $\{y_{p,\tau}\}_{\tau=1}^{T}$, to model their temporal correlations. The output of the temporal Transformer is further processed using a feed-forward network (FFN), and then integrated through a 3D convolution layer to obtain the spatio-temporal visual feature \(\tilde{M} \in \mathbb{R}^{C \times h \times w}\). This method is the one we ultimately choose.

\vspace{-0.2cm}\subsection{Audio Feature Representation}\vspace{-0.1cm}
The auditory input for OmniAVS is the 2-second first-order ambisonics audio fragment corresponding to the segmented video clip. The raw waveforms of the four ambisonics channels are resampled to 16kHz and then converted into 2D Mel spectrograms using 128 mel-spectrogram bins, capturing both the temporal and spectral characteristics of the audio signals. The Mel spectrogram represents audio in a way that aligns with human auditory perception, with frequency bands logarithmically spaced to reflect perceptually significant ranges. The resulting Mel spectrogram, a 2D representation akin to an image, is well-suited for processing by vision-based models. In OmniAVS, the spectrogram is fed into ImageBind’s pre-trained ViT-B/16 audio encoder, which has been trained to extract high-level semantic features from the spectrograms. This encoder projects the audio into a joint embedding space aligned with visual semantics. This alignment enables the model to effectively capture how different auditory cues influence visual attention, facilitating more accurate saliency predictions. Additionally, multi-level features are extracted from the audio encoder at different stages. For the \(j\)-th encoder stage, the extracted features are denoted as \(A_j^{\text{W}}, A_j^{\text{X}}, A_j^{\text{Y}}, A_j^{\text{Z}} \). This multi-level feature extraction enables the model to capture both fine-grained and abstract representations of the audio, further enriching the fusion of auditory and visual information.


\vspace{-0.1cm}\subsection{Audio-Visual Feature Fusion}
\begin{figure}[t!]
\setlength{\abovecaptionskip}{0cm}
\setlength{\belowcaptionskip}{-0.2cm}
\vspace{-0.1cm}
\centering
\includegraphics[width=1\linewidth]{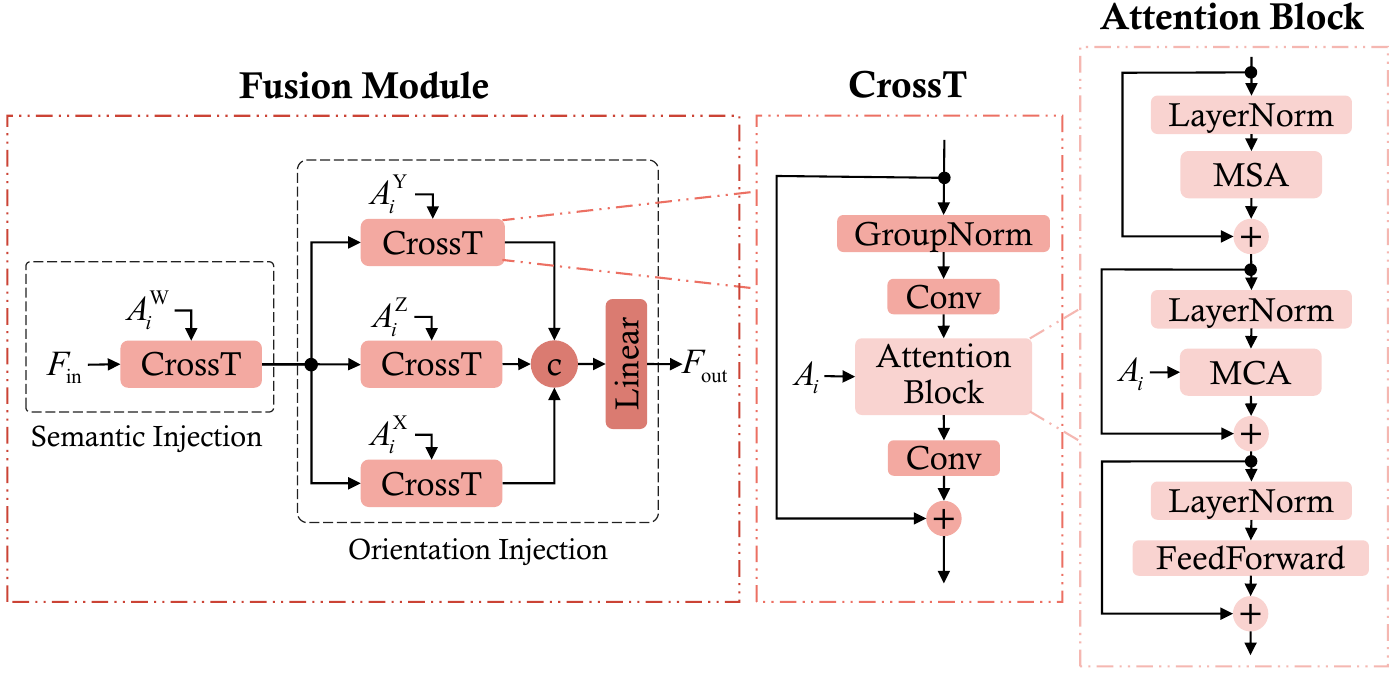}
\caption{The framework of the feature fusion module. ``CrossT" stands for CrossTransformer. ``MSA'' represents multi-head self-attention. ``MCA'' indicates multi-head cross-attention.
}
\label{fig:fusion}
\vspace{-0.3cm}
\end{figure}

Inspired by the approach of infusing textual features into visual features  used in Stable Diffusion \cite{9878449}, we develop an audio-visual feature fusion module to progressively inject multi-level audio features into the visual features via cross-attention. The framework of the feature fusion module is shown in Fig. \ref{fig:fusion}, which is divided into two stages, including a semantic injection stage and an orientation injection stage. The audio feature corresponding to the W channel, $A^{\text{W}}$, which primarily offers semantic information, is first injected into the input feature $F_\text{in}$ through a CrossTransformer:
\begin{equation}
F_{\text{av}}^{\text{W}}=\mathrm{CrossTransformer}(F_{\text{in}},A^{\text{W}}).
\end{equation}
Specifically, the CrossTransformer first performs multi-head self-attention (MSA) on the input feature \( F_{\text{in}} \), resulting in \( F' \). This output is then projected to form the ``query'', while the ``key'' and ``value'' are derived from $A^{\text{W}}$, to perform multi-head cross-attention (MCA).
The output $F^{''}$ are then fed into a FeedForward module to output the fused audio-visual feature $F^{\text{W}}_{\text{av}}$. The CrossTransformer is formulated as:\vspace{-0.05cm}
\begin{equation}
    F' = \mathrm{MSA}\left(\mathrm{LN}\left(F_{\text{in}}\right)\right)+F_{\text{in}},\vspace{-0.05cm}
\end{equation}
\begin{equation}
F''=\mathrm{MCA}\left(\mathrm{LN}\left(F'\right),A^{\text{W}}\right)+F',\vspace{-0.05cm}
\end{equation}
\begin{equation}
    F_{\text{av}}^{\text{W}} = \mathrm{FeedForward} \left(\mathrm{LN}\left(F''\right)\right)+F''.\vspace{-0.05cm}
\end{equation}
In the orientation injection stage, $A^{\text{X}}, A^{\text{Y}}, A^{\text{Z}}$ are injected into $F_{\text{av}}^{\text{W}}$ using the same way, respectively as follows. 
\begin{equation}
F_{\text{av}}^{\text{Y}}=\mathrm{CrossTransformer}(F_{\text{av}}^{\text{W}},A^{\text{Y}}).\vspace{-0.05cm}
\end{equation}
\begin{equation}
F_{\text{av}}^{\text{Z}}=\mathrm{CrossTransformer}(F_{\text{av}}^{\text{W}},A^{\text{Z}}).\vspace{-0.05cm}
\end{equation}
\begin{equation}
F_{\text{av}}^{\text{X}}=\mathrm{CrossTransformer}(F_{\text{av}}^{\text{W}},A^{\text{X}}).\vspace{-0.05cm}
\end{equation}
The fused audio-visual features are then concatenated, and a linear layer is finally applied to integrate information from all three directions:\vspace{-0.05cm}
\begin{equation}\vspace{-0.05cm}
F_{\text{out}}=\mathrm{Linear}\left(\mathrm{Concat}\left(F_{\text{av}}^{\text{X}}, F_{\text{av}}^{\text{Y}}, F_{\text{av}}^{\text{Z}}\right)\right).
\end{equation}

\subsection{Audio-Visual Saliency Estimation}
Audio-Visual Saliency Estimation Blocks progressively decode the audio-visual features to compute the saliency of the video.
In each block, the audio-visual features are first upsampled using the up-sampling components constituting with a convolution and a pixelshuffle layer, and subsequently merged with the features from the visual encoder via skip connections and $1\times1$ convolution layer, ensuring the preservation and integration of high-resolution features from the encoder. 

\subsection{Saliency Losses}
When measuring the error between the predicted saliency maps $S_{\text{pd}}$ and ground truth saliency maps $S_{\text{gt}}$, our method incorporates multiple loss functions, ensuring a comprehensive and effective training process for optimizing the network parameters $\mathbf{W}$.

The L1 loss computes the sum of the absolute differences between the predicted and ground truth saliency maps:\vspace{-0.15cm}
\begin{equation}
    \mathcal{L}_{\text{L1}}(\mathbf{W}) =\sum_{i,j}^{N_{x,y}} \left|S_{\text{gt}}(i,j|\mathbf{W})-S_{\text{pd}}(i,j|\mathbf{W})\right| .\vspace{-0.15cm}
\end{equation}
The Correlation Coefficient (CC) Loss measures the linear correlation between the predicted and ground truth maps:
\begin{equation}
    \mathcal{L}_{\text{CC}}(\mathbf{W}) =-\frac{\mathrm{cov}(S_{\text{pd}}(x,y|\mathbf{W}),S_{\text{gt}}(x,y|\mathbf{W}))}{\sigma{(S_{\text{pd}}(x,y|\mathbf{W}))} \sigma(S_{\text{gt}}(x,y|\mathbf{W}))}.
\end{equation}
The Kullback-Leibler (KL) Divergence Loss assesses the probabilistic distribution difference between the predicted and ground truth saliency maps: \vspace{-0.15cm}
 \begin{equation}
     \mathcal{L}_{\text{KL}}(\mathbf{W}) = \sum_{i,j}^{N_{x,y}} S_{\text{gt}}(i,j|\mathbf{W})\ln{\left(\varepsilon+\frac{S_{\text{gt}}(i,j|\mathbf{W})}{\varepsilon+S_{\text{pd}}(i,j|\mathbf{W})}\right)}.\vspace{-0.15cm}
 \end{equation}
The Binary Cross-Entropy (BCE) Loss, a standard choice for binary classification tasks, is also applied to further refine the model's ability to discern salient from non-salient regions:
 \begin{equation}\hspace{-0.5cm}
 \begin{split}
     \mathcal{L}_{\text{BCE}}(\mathbf{W}) &= -\frac{1}{N_{x,y}} \sum_{i,j}^{N_{x,y}} [S_{\text{gt}}(i,j|\mathbf{W}) \cdot \log(S_{\text{pd}}(i,j|\mathbf{W})) \\
     \vspace{-0.5cm} &+ (1 - S_{\text{gt}}(i,j|\mathbf{W})) \cdot \log(1 - S_{\text{pd}}(i,j|\mathbf{W}))].
 \end{split}\vspace{-0.15cm}
 \end{equation}
 The final loss can be derived as a weighted sum of these diverse loss functions:
 \begin{equation}
     \mathcal{L} = \omega_1 \mathcal{L}_{\text{L1}}+\omega_2 \mathcal{L}_{\text{CC}}+\omega_3 \mathcal{L}_{\text{KL}}+\omega_4 \mathcal{L}_{\text{BCE}}.
 \end{equation}

\section{Experiments}\label{experiments}
\subsection{Experimental Setup}\label{sec:dataset}
\subsubsection{Databases}
We first validate the overall effectiveness of OmniAVS on three audio-visual ODV saliency datasets: AVS-ODV, YQ-ODV \cite{10224292}, and LJ-ODV \cite{9897737}. For AVS-ODV, we randomly select 30 videos as testing data, and use the remaining videos as the training data.
The training/testing split operation is repeated three times to obtain three splits. The final results are calculated as the average of the performance scores across the three splits. For YQ-ODV \cite{10224292}, we adopt the standard division provided by this dataset, which consists of 46 training videos and 11 testing videos. For LJ-ODV, 43 videos are randomly divided into 30 training and 13 testing videos. Moreover, we further assess the effectiveness of the visual part of OmniAVS using two visual-only saliency datasets for ODVs, PVS-HM \cite{8418756} and Zhang-ODV \cite{Zhang_2018_ECCV}. For PVS-HM  \cite{8418756}, we follow the setting provided by the dataset and divide 76 videos into 61 training videos and 15 testing videos. For Zhang-ODV \cite{Zhang_2018_ECCV}, we follow the setting provided by the dataset and divide 104 videos into 80 training videos and 24 test videos. Additionally, we evaluate the generalizability of OmniAVS using six traditional video audio-visual saliency datasets, including DIEM \cite{Mital2011ClusteringOG}, Coutrot1 \cite{Coutrot2014HowSF}, Coutrot2 \cite{Coutrot2016}, AVAD \cite{7457921}, SumMe \cite{Gygli2014CreatingSF}, and ETMD \cite{10.1016/j.image.2015.08.004}, where the training is performed on the combined dataset from all six. For DIEM, we use the standard division \cite{6253254} for testing. For the other five datasets, we adopt the three-fold cross-validation division method used by Tsiami \textit{et al.} \cite{tsiami2020stavis}.

\subsubsection{Implementation Details}
We implement the OmniAVS network using the PyTorch framework. The weights of ImageBind are frozen. The other parts of the network is optimized using SGD with an initial learning rate of 0.1, governed by a cosine annealing schedule with warm-restarts. The hyperparameters in the loss function, \( \omega_1 \), \( \omega_2 \), \( \omega_3 \), and \( \omega_4 \), are empirically set to 1, 0.2, 0.2, and 1, respectively.

To assess the saliency prediction model, we adopt six mainstream evaluation metrics, including NSS, SIM, CC, AUC-Judd (AUC-J) \cite{Judd}, shuffled AUC (s-AUC \cite{10.1167/7.14.4}), and Kullback-Leibler divergence (KLD). The KLD can be calculated in two ways depending on the choice of ground truth. In most cases, we use the fixation map as the reference to calculate the KLD metric, while for evaluating YQ-ODV and PVS-HM datasets, the saliency map is used. 
We also employ the sinusoids weighting method described in Section \ref{subsec:sim} to correct the geometric distortions introduced by the ERP format.

\vspace{-0.1cm}\subsection{Comparison with State-of-the-Arts on Audio-Visual ODV Saliency Databases}
\useunder{\uline}{\ul}{}
\definecolor{mygray}{gray}{0.9}
\begin{table*}[ht]
\setlength{\abovecaptionskip}{0cm}
\setlength{\belowcaptionskip}{0cm}
\vspace{-0.5cm}
\setlength\tabcolsep{5pt} 
\centering
\caption{Performance comparison of different saliency prediction models on the AVS-ODV database. ``FT" denotes fine-tuning on AVS-ODV based on pre-trained model weights. ``w/o p" and ``w/ p" indicate without and with pre-training on traditional audio-visual datasets, respectively. In each category of models, the best results for each metric are highlighted in \textbf{bold}, and the second-best results are {\ul underlined}. The best results throughout the table for each metric are shown in  {\color[HTML]{FF0000} red}, while the second-best results are shown in  {\color[HTML]{0070C0} blue}. The same rules apply to the following tables as well.\label{tab:avsodv}}
\begin{adjustbox}{width=0.94\textwidth}
\begin{tabular}{l|cccccc|cccccc|cccccc}
\toprule
Audio Mode & \multicolumn{6}{c|}{Ambisonics} & \multicolumn{6}{c|}{Mono} & \multicolumn{6}{c}{Mute} \\ \midrule
Model\textbackslash{}Metric & {\scriptsize NSS↑} & {\scriptsize SIM↑}& {\scriptsize CC↑} &{\scriptsize AUC-J↑} & {\scriptsize s-AUC}↑ & {\scriptsize KLD↓} &{\scriptsize NSS↑} &{\scriptsize SIM↑} & {\scriptsize CC↑} &{\scriptsize AUC-J↑} &{\scriptsize s-AUC↑} & {\scriptsize KLD↓} & {\scriptsize NSS↑} & {\scriptsize SIM↑}& {\scriptsize CC↑} & {\scriptsize AUC-J↑} &{\scriptsize s-AUC↑} & {\scriptsize KLD↓} \\ \midrule
IT \cite{730558}& 0.6281 & 0.1372 & 0.1090 & 0.6512 & 0.5034 & 22.937 & 0.6026 & 0.1383 & 0.1082 & 0.6491 & 0.5017 & 23.031 & 0.5713 & 0.1537 & 0.1199 & 0.6412 & 0.4963 & 23.410 \\
GBVS \cite{10.5555/2976456.2976525} & 1.2738 & {\ul 0.1992} & 0.2298 & 0.8293 & 0.5343 & 8.8657 & 1.2961 & {\ul 0.2094} & 0.2398 & 0.8333 & 0.5408 & 8.8424 & 1.2805 & {\ul 0.2447} & 0.2754 & 0.8326 & 0.5331 & 8.8528 \\
SR \cite{SR}& 1.0485 & 0.1770 & 0.1852 & 0.8070 & 0.5482 & 8.9671 & 1.0341 & 0.1833 & 0.1874 & 0.8059 & 0.5467 & 8.9737 & 0.9604 & 0.2130 & 0.2045 & 0.7938 & 0.5245 & 9.0277 \\
SMVJ \cite{SMVJ}& 1.4023 & \textbf{0.2064} & {\ul 0.2509} & 0.8442 & 0.5355 & {\ul 8.7974} & 1.4266 & \textbf{0.2170} & 0.2617 & 0.8481 & 0.5429 & {\ul 8.7745} & 1.3918 & \textbf{0.2524} & {\ul 0.2975} & 0.8459 & 0.5334 & {\ul 8.7955} \\
SUN \cite{SUN}& \textbf{1.6730} & 0.1907 & 0.2436 & {\ul 0.8773} & \textbf{0.5860} & \textbf{8.6761} & \textbf{1.6781} & 0.1976 & 0.2516 & {\ul 0.8772} & \textbf{0.5869} & \textbf{8.6767} & \textbf{1.5875} & 0.2308 & 0.2806 & 0.8667 & \textbf{0.5642} & \textbf{8.7181} \\
PQFT \cite{PQFT}& 0.8889 & 0.1702 & 0.1556 & 0.7926 & 0.5408 & 9.0414 & 0.8785 & 0.1761 & 0.1580 & 0.7921 & 0.5400 & 9.0436 & 0.8283 & 0.2062 & 0.1751 & 0.7830 & 0.5237 & 9.0813 \\
Judd \cite{Judd}& 1.3816 & 0.1798 & 0.2463 & 0.8618 & 0.5371 & 8.8955 & 1.4299 & 0.1893 & {\ul 0.2619} & 0.8692 & 0.5538 & 8.8767 & 1.4168 & 0.2230 & \textbf{0.3023} & {\ul 0.8673} & 0.5503 & 8.8798 \\
SWD \cite{SWD}& 1.2979 & 0.1870 & 0.2243 & 0.8189 & 0.5369 & 8.9007 & 1.3721 & 0.1993 & 0.2435 & 0.8284 & 0.5544 & 8.8631 & 1.3841 & 0.2351 & 0.2848 & 0.8328 & {\ul 0.5569} & 8.8516 \\
Murray \cite{Murray}& {\ul 1.5121} & 0.1833 & \textbf{0.2575} & \textbf{0.8838} & {\ul 0.5769} & 8.8196 & {\ul 1.5032} & 0.1898 & \textbf{0.2640} & \textbf{0.8817} & {\ul 0.5714} & 8.8240 & {\ul 1.4217} & 0.2224 & 0.2946 & \textbf{0.8682} & 0.5416 & 8.8553 \\
HFT \cite{HFT}& 1.0375 & 0.1929 & 0.1895 & 0.8062 & 0.5222 & 8.9785 & 1.0707 & 0.2032 & 0.1996 & 0.8115 & 0.5342 & 8.9482 & 1.0698 & 0.2385 & 0.2292 & 0.8132 & 0.5348 & 8.9463 \\ \midrule
SALICON \cite{7410395}& 2.1414 & 0.2509 & 0.3613 & 0.8909 & {0.5692} & 8.4353 & 2.1105 & 0.2589 & 0.3665 & 0.8905 & 0.5665 & 8.4407 & 1.8789 & 0.2905 & 0.3856 & 0.8754 & 0.5302 & {\ul 8.5557} \\
MLNet \cite{mlnet}& {\ul 2.3477} & 0.2564 &{\ul0.3807} & 0.8904 & \uline{0.5772} & 8.6060 & {2.3451} & 0.2659 & \uline{0.3901} & 0.8898 & \uline{0.5765} & 8.6333 & {\ul 2.0441} & 0.2963 & {\ul 0.4022} & 0.8741 & 0.5387 & 8.7560 \\
SalGAN \cite{Pan_2017_SalGAN}& 2.1168 & 0.2638 & 0.3626 & \uline{0.9042} & 0.5620 & {8.3953} & 2.1090 & 0.2738 & 0.3716 & \uline{0.9051} & 0.5637 & \uline{8.3828} & 1.9433 & 0.3089 & \textbf{0.4025} & \textbf{0.8917} & 0.5307 & \textbf{8.5063} \\
SAM \cite{SAM}& 2.2757 & 0.2849 & 0.3733 & {0.9031} & 0.5652 & 8.3873 & 2.2587 & {0.2952} & 0.3814 & {0.9033} & {0.5668} & {8.3843} & 2.0157 & {\ul 0.3263} & 0.4006 & 0.8875& 0.5298 & 8.5872 \\
VQSal \cite{duan2022saliency}& 2.0596 & 0.2678 & 0.3517 & 0.8985 & 0.5549 & 8.5562 & 2.0696 & 0.2797 & 0.3634 & 0.8999 & 0.5595 & 8.5680 & 1.9129 & 0.3141 & 0.3932 & {\ul 0.8889} & 0.5299 & 8.7565\\
TASED \cite{min2019tased}& 2.2755 & {\ul0.3040} & 0.3644 & 0.8319 & 0.5516 & 12.213 & \uline{2.3554} & \uline{0.3213} & {0.3849} & 0.8377 & 0.5609 & 12.027 & \textbf{2.0820} & \textbf{0.3411} & 0.4007 & 0.8269 & {\ul 0.5393} & 12.481 \\
SalEMA \cite{salema}& 1.8455 & 0.2629 & 0.3000 & 0.8460 & 0.5382 & 9.5731 & 1.9335 & 0.2811 & 0.3197 & 0.8522 & 0.5516 & 9.4825 & 1.8077 & 0.3170 & 0.3535 & 0.8491 & \textbf{0.5431} & 9.6605 \\
STAViS \cite{tsiami2020stavis}& 1.9652 & 0.2714 & 0.3272 & 0.8645 & 0.5499 & 8.9092 & 2.0090 & 0.2872 & 0.3444 & 0.8694 & 0.5586 & 8.8501 & 1.6717 & 0.3072 & 0.3416 & 0.8411 & 0.5349 & 9.3504 \\
DAVE \cite{dave}& 1.9095 & 0.2663 & 0.3142 & 0.8629 & 0.5427 & 9.3329 & 2.0098 & 0.2873 & 0.3389 & 0.8707 & 0.5587 & 9.1963 & - & - & - & - & - & - \\
{DiffSal \cite{xiong2024diffsal}}&{\textbf{3.1606}}&{\textbf{0.3697}}&{\textbf{0.4982}}&{\textbf{0.9182}}&{\textbf{0.5896}}&{\textbf{8.1804}}&{\textbf{3.1416}}&{\textbf{0.3815}}&{\textbf{0.5081}}&{\textbf{0.9185}}&{\textbf{0.5892}}&{\textbf{8.1891}}&{-}&{-}&{-}&{-}&{-}&{-}\\
AVS360 \cite{avs360}& 1.4197 & 0.2401 & 0.2447 & 0.7783 & 0.5283 & 13.881 & 1.5661 & 0.2660 & 0.2765 & 0.8106 & 0.5475 & 12.342 & - & - & - & - & - & - \\
SVGC-AVA \cite{10224292}& 1.5341 & 0.2582 & 0.2568 & 0.9022 & 0.5329 & 8.4721 & - & - & - & - & - & - & - & - & - & - & - & - \\ \midrule
SALICON(FT) \cite{7410395} & 3.5472 & 0.3839 & 0.5677 & {0.9459} & {0.6107} & {7.8249} & 3.5072 & 0.3895 & 0.5769 & \uline{0.9464} & {0.6113} & {7.8293} & 2.8552 & 0.3953 & 0.5627 & 0.9285 & 0.5445 & {\ul 8.0576} \\
MLNet(FT) \cite{mlnet}& {4.2095} & 0.4141 & 0.5594 & 0.8788 & 0.6055 & 12.195 & {4.1620} & 0.4227 & 0.5672 & 0.8903 & 0.6007 & 11.422 & 3.2497 & 0.4148 & 0.5401 & 0.8946 & 0.5505 & 10.360 \\
SalGAN(FT) \cite{Pan_2017_SalGAN}& 3.9515 & {0.4388} & {0.5723} & 0.9438 & 0.6080 & 8.2416 & 3.9434 & {0.4488} & 0.5756 & 0.9432 & 0.6021 & 8.3639 & 3.2183 & {\ul 0.4494} & 0.5732 & {\ul 0.9297} & {\ul 0.5561} & 8.4370 \\
SAM(FT) \cite{SAM}& 3.4284 & 0.3480 & 0.4813 & 0.9285 & 0.5973 & 8.2865 & 3.6236 & 0.3731 & 0.4818 & 0.9228 & 0.5941 & 8.9512 & 2.7811 & 0.3590 & 0.4539 & 0.8788 & 0.5437 & 10.867 \\
VQSal(FT) \cite{duan2022saliency} & 4.1704 & 0.4306 & 0.5527 & 0.9370 & 0.6030 & 8.4798 & \uline{4.1829} & 0.4482 & 0.5618 & 0.9398 & 0.5962 & 8.4238 & \textbf{3.4543} & 0.4483 & 0.5603 & 0.9269 & {\ul 0.5590} & 8.6498\\
TASED(FT) \cite{min2019tased}& \uline{4.2638} & \uline{0.4647} & \uline{0.6114} & 0.9352 & \uline{0.6174} & 8.9680 & 4.1074& \uline{0.4675} & \uline{0.6094} & 0.9419 & \uline{0.6120} & 8.4524 & {\ul3.3827} & \textbf{0.4634} & \textbf{0.6075} & 0.9261 & \textbf{0.5600} & 8.8411 \\
SalEMA(FT) \cite{salema}& 3.9582 & 0.4263 & 0.5708 & \uline{0.9466} & 0.6020 &{\uline{7.6986}} & 3.9226 & 0.4373 & {0.5776} & {0.9460} & 0.5951 & \uline{7.7543} & 3.2061 & 0.4365 & {\ul 0.5757} & \textbf{0.9312} & 0.5455 & \textbf{7.9189} \\
STAViS(FT) \cite{tsiami2020stavis}& 3.5850 & 0.4171 & 0.5468 & 0.9406 & 0.5962 & 8.0112 & 3.5289 & 0.4257 & 0.5525 & 0.9408 & 0.5923 & 7.9894 & 2.3730 & 0.3873 & 0.4739 & 0.9110 & 0.5260 & 8.3350 \\
DAVE(FT) \cite{dave}& 3.1711 & 0.3839 & 0.4981 & 0.8749 & 0.5952 & 11.764 & 3.1466 & 0.3916 & 0.5059 & 0.8752 & 0.5899 & 11.788 & - & - & - & - & - & - \\
	
{DiffSal(FT) \cite{xiong2024diffsal}}&{\textbf{4.8315}}&{\textbf{0.4832}}&{\color[HTML]{FF0000}\textbf{0.6696}}&{\color[HTML]{FF0000}\textbf{0.9572}}&{\textbf{0.6300}}&{\color[HTML]{FF0000}\textbf{7.4027}}&{\textbf{4.8026}}&{\textbf{0.4855}}&{\color[HTML]{0070C0} \textbf{0.6706}}&{\color[HTML]{FF0000}\textbf{0.9562}}&{\textbf{0.6290}}&{\color[HTML]{FF0000}\textbf{7.4735}}&{-}&{-}&{-}&{-}&{-}&{-}\\ 	 	 	 	 	 	 	 	 	
AVS360(FT) \cite{avs360}& 3.1209 & 0.3789 & 0.4665 & 0.9209 & 0.5732 & 8.7508 & 3.1346 & 0.3942 & 0.4828 & 0.9270 & 0.5824 & 8.6060 & - & - & - & - & - & - \\
SVGC-AVA(FT) \cite{10224292}& 2.0928 & 0.2869 & 0.3507 & 0.9169 & 0.5421 & 8.2259 & - & - & - & - & - & - & - & - & - & - & - & - \\ \midrule
\rowcolor{mygray}
OmniAVS(w/o p) & {\color[HTML]{0070C0} {\ul 5.0703}} & {\color[HTML]{0070C0} {\ul 0.4838}} & {{\ul 0.6452}} & {{\ul 0.9541}} & {\color[HTML]{0070C0}{\ul 0.6357}} & {\ul 7.7197} & {\color[HTML]{0070C0} {\ul 5.0972}} & {\color[HTML]{0070C0} {\ul 0.4984}} & {{\ul 0.6639}} & {{\ul 0.9551}} & {\color[HTML]{0070C0} {\ul 0.6367}} & {{\ul 7.6568}} & {\color[HTML]{0070C0} {\ul 4.0125}} & {\color[HTML]{0070C0} {\ul 0.4738}} & {\color[HTML]{0070C0} {\ul 0.6350}} & {\color[HTML]{0070C0} {\ul 0.9403}} & {\color[HTML]{0070C0} {\ul 0.5734}} & {\color[HTML]{FF0000} \textbf{7.8898}} \\ 
\rowcolor{mygray}
OmniAVS(w/ p) & {\color[HTML]{FF0000} \textbf{5.2461}} & {\color[HTML]{FF0000} \textbf{0.5003}} & {\color[HTML]{0070C0} \textbf{0.6662}} & {\color[HTML]{0070C0} \textbf{0.9557}} & {\color[HTML]{FF0000} \textbf{0.6437}} & {\color[HTML]{0070C0}\textbf{7.6665}} & {\color[HTML]{FF0000} \textbf{5.2221}} & {\color[HTML]{FF0000} \textbf{0.5086}} & {\color[HTML]{FF0000} \textbf{0.6786}} & {\color[HTML]{0070C0} \textbf{0.9560}} & {\color[HTML]{FF0000} \textbf{0.6396}} & {\color[HTML]{0070C0} \textbf{7.6294}} & {\color[HTML]{FF0000} \textbf{4.0588}} & {\color[HTML]{FF0000} \textbf{0.4823}} & {\color[HTML]{FF0000} \textbf{0.6440}} & {\color[HTML]{FF0000} \textbf{0.9408}} & {\color[HTML]{FF0000} \textbf{0.5749}} & {\color[HTML]{0070C0} {\ul 7.9042}} \\ \bottomrule
\end{tabular}
\end{adjustbox}
\vspace{-0.3cm}
\end{table*}

We first conduct experiments on AVS-ODV to validate the effectiveness of OmniAVS in saliency prediction for ODVs. We compare the performance of OmniAVS with ten classical saliency models, including IT \cite{730558}, GBVS \cite{10.5555/2976456.2976525}, SR \cite{SR}, SMVJ \cite{SMVJ}, SUN \cite{SUN}, PQFT \cite{PQFT}, Judd \cite{Judd}, SWD \cite{SWD}, Murray \cite{Murray}, HFT \cite{HFT}, and twelve state-of-the-art deep learning-based models, including SALICON \cite{7410395}, MLNet \cite{mlnet}, SalGAN \cite{Pan_2017_SalGAN}, SAM \cite{SAM},  VQSal \cite{duan2022saliency}, TASED \cite{min2019tased}, SalEMA \cite{salema}, STAViS \cite{tsiami2020stavis}, DAVE \cite{dave}, DiffSal \cite{xiong2024diffsal}, AVS360 \cite{avs360}, and SVGC-AVA \cite{10224292}. For deep learning models, we assess their performance using both their official pre-trained weights and the models fine-tuned on our AVS-ODV database. For OmniAVS, we evaluate its performance after training only on AVS-ODV database, as well as its performance after pre-training on traditional audio-visual saliency datasets described in Section \ref{sec:dataset} followed by training on AVS-ODV database. We conduct experiments for each audio mode in AVS-ODV database. In terms of our OmniAVS net, we remove the orientation injection part of the fusion module for the mono mode, and remove the entire fusion module for the mute mode.

Table \ref{tab:avsodv} provides a quantitative comparison of the performance between different models. OmniAVS significantly outperforms all other models, especially after pre-training on traditional audio-visual datasets. Models based on traditional handcrafted features demonstrate limited performance. Deep learning-based models show improvement in performance after fine-tuning on the AVS-ODV database. Among them, TASED \cite{min2019tased} and SalEMA \cite{salema}, which are visual-only saliency prediction models for traditional videos, achieve commendable performance. DiffSal \cite{xiong2024diffsal}, an audio-visual saliency prediction model for traditional videos, achieves the second-highest performance after OmniAVS, benefiting from its diffusion-based model, which iteratively refines saliency maps, and its use of cross-attention mechanisms for effective audio-visual integration. However, STAViS \cite{tsiami2020stavis} and DAVE \cite{dave}, despite integrating audio, do not perform as well, suggesting that the visual parts of these two methods need to be improved. AVS360 \cite{avs360} and SVGC-AVA \cite{10224292} are networks specifically designed for audio-visual ODV saliency prediction but demonstrate relatively limited performance. The poor performance of SVGC-AVA can be attributed to its exclusive use of audio energy distribution as the audio input, without considering the semantic information of audio in complex audio-visual scenes, and is also due to the model's lack of consideration for temporal information.
\begin{figure*}[t!]
\centering
\subfloat[Type 1]{\includegraphics[height=0.6\linewidth]{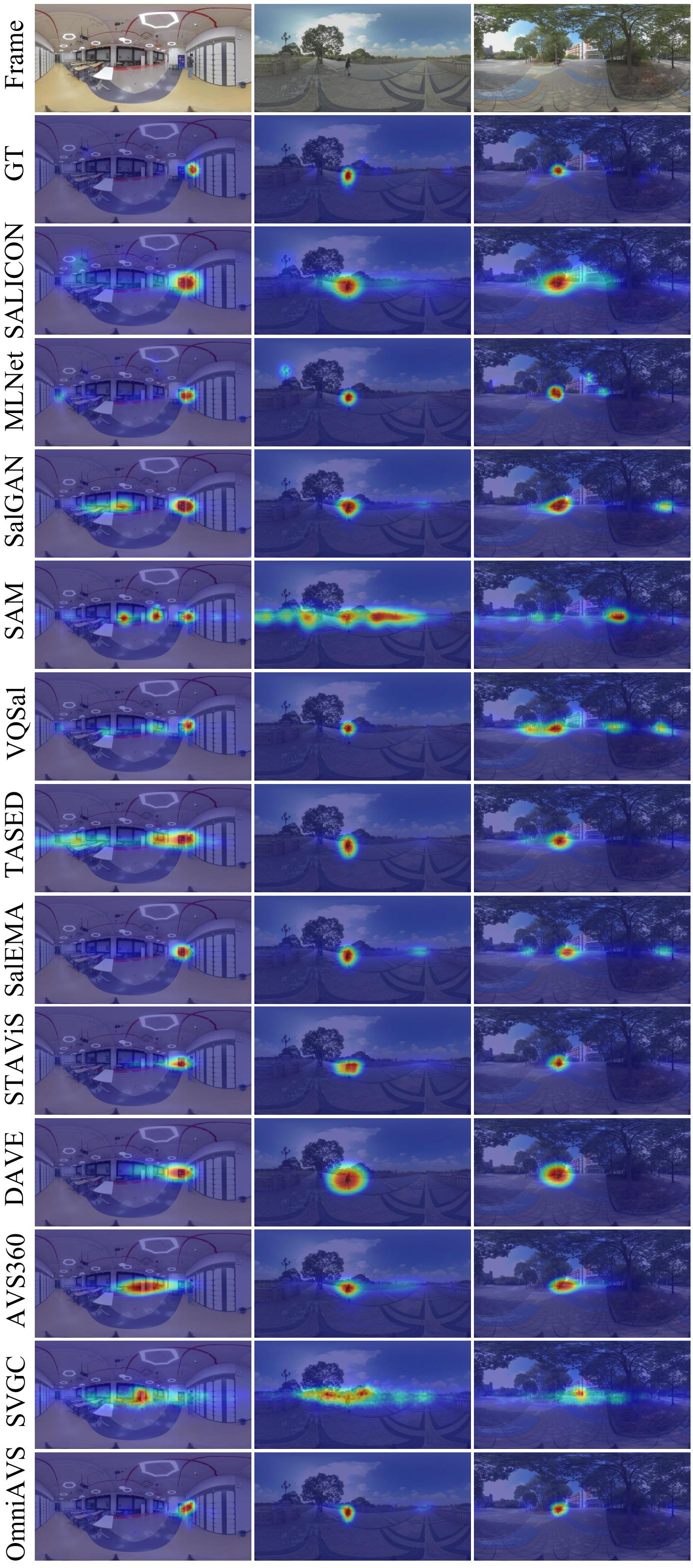}} \hspace{0.02pt}
\subfloat[Type 2]{\includegraphics[height=0.6\linewidth]{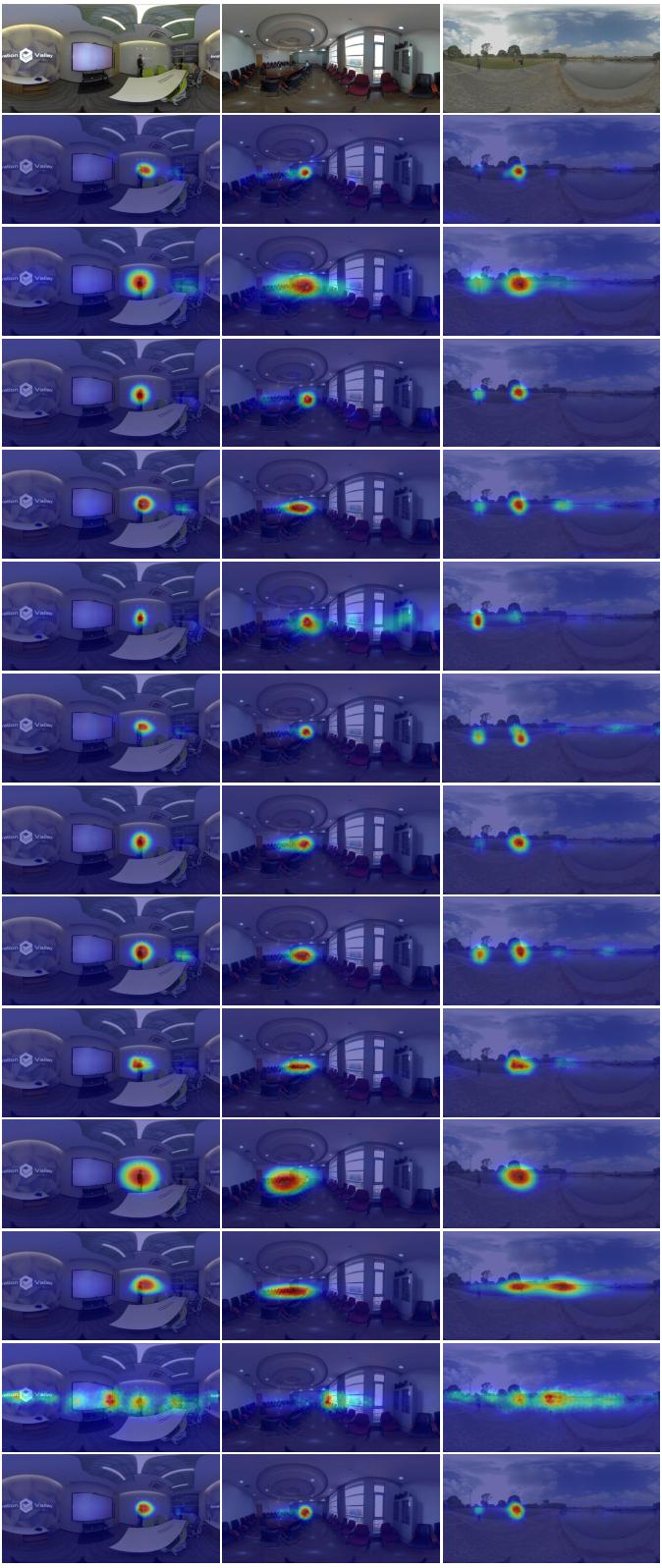}}\hspace{0.02pt}
\subfloat[Type 3]{\includegraphics[height=0.6\linewidth]{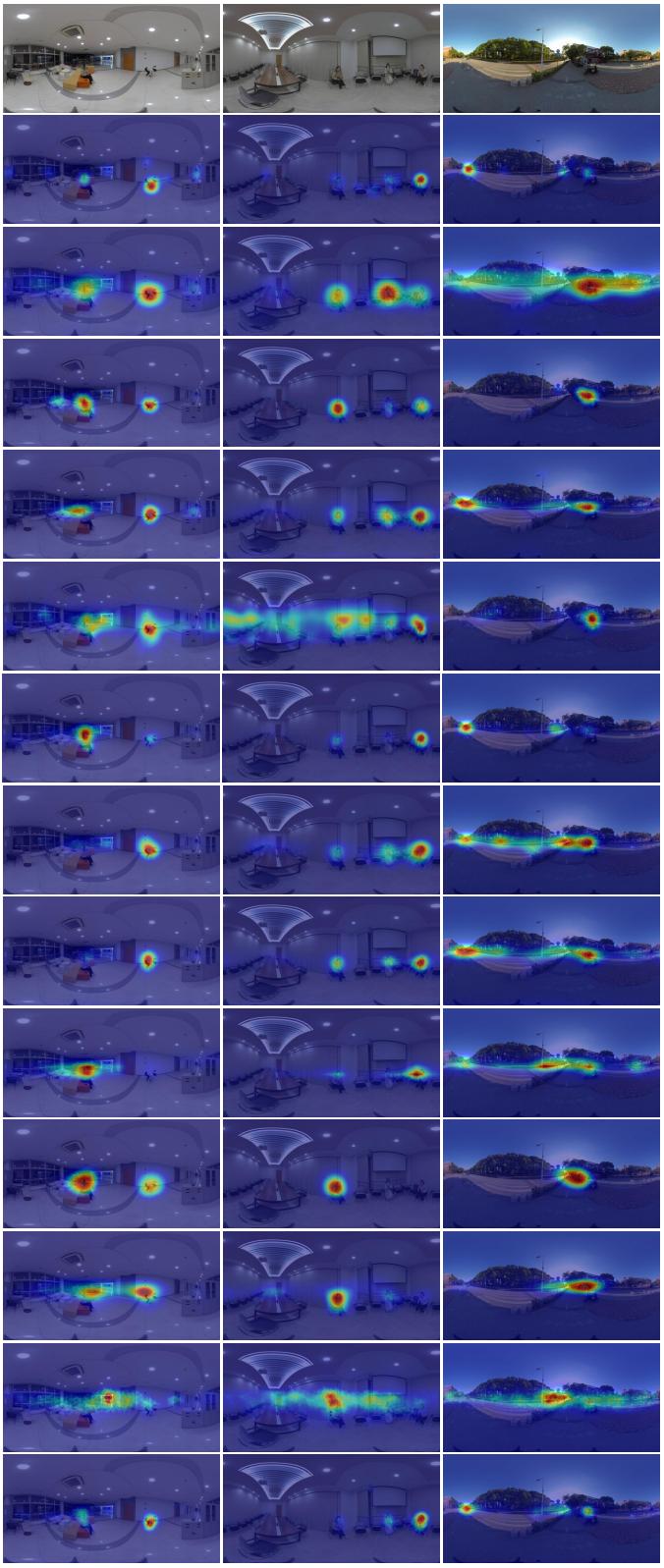}}
\caption{Qualitative comparison of the performance of OmniAVS and state-of-the-art saliency models on three types of videos in AVS-ODV database.}
\label{fig:pred}
\vspace{-0.3cm}
\end{figure*}
\begin{figure}[t!]
\vspace{-0.1cm}
\setlength{\abovecaptionskip}{0cm}
\centering
\includegraphics[width=3.45in]{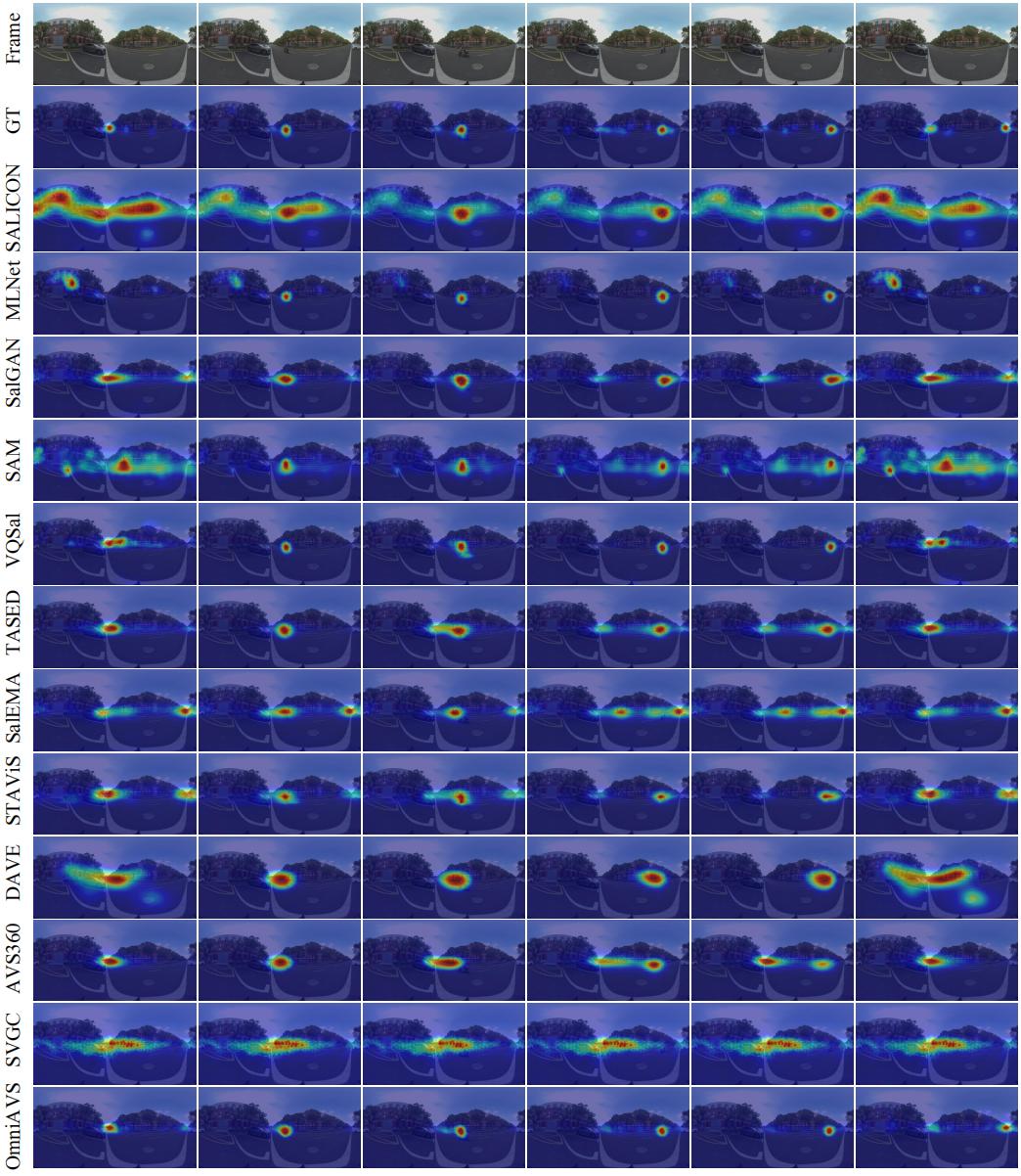}
\caption{Qualitative comparison of the performance of OmniAVS and state-of-the-art saliency models on consecutive multiple video frames of a video from the AVS-ODV database.}
\label{fig:predd}
\vspace{-0.4cm}
\end{figure}

Furthermore, we compare the perceived performance of OmniAVS with the fine-tuned versions of all state-of-the-art DNN-based methods on AVS-ODV. Fig. \ref{fig:pred} illustrates the video frames of nine ODVs from our AVS-ODV database, along with their corresponding ground truth and predicted saliency maps under the ambisonics mode. Fig. \ref{fig:predd} shows the results of multiple frames of one video. The saliency maps predicted by OmniAVS exhibit the highest alignment with the ground truth, demonstrating OmniAVS's ability to effectively integrate visual and auditory information to produce more accurate saliency maps.

We further evaluate OmniAVS on two more audio-visual ODV sailency databases. Table \ref{tab:yqodv} presents the performance of each model under the ambisonics audio mode on YQ-ODV \cite{10224292}, and Table \ref{tab:ljodv} displays the performance of different models under the stereo and mute audio modes on LJ-ODV \cite{9897737}. All the results demonstrate the superior performance of the OmniAVS net on the task of omnidirectional audio-visual saliency prediction.

\begin{table}[t!]
\setlength\tabcolsep{5pt} 
\setlength{\abovecaptionskip}{0cm}
\caption{Performance comparison of different saliency prediction models on the YQ-ODV database. 
\label{tab:yqodv}}
\centering
\begin{adjustbox}{width=0.34\textwidth}
\begin{tabular}{l|ccccc}
\toprule
Model\textbackslash{}Metric & NSS↑ & SIM↑ & CC↑ & {\scriptsize AUC-J}↑ & KLD↓ \\ \midrule
DAVE \cite{dave}& 3.398 & {\ul 0.503} & 0.676 & 0.926 & 1.546 \\
STAViS \cite{tsiami2020stavis}& 1.934 & 0.321 & 0.432 & 0.892 & 1.525 \\
CP360 \cite{8578252}& 1.492 & 0.205 & 0.328 & 0.883 & 2.026 \\
SST-Sal \cite{BERNALBERDUN2022200}& 2.073 & 0.348 & 0.429 & 0.903 & 1.537 \\
AVS360 \cite{avs360}& 3.242 & 0.471 & 0.659 & 0.916 & 1.334 \\
SVGC-AVA \cite{10224292}& {\ul 3.951} & 0.498 & {\ul 0.719} & {\ul 0.938} & \textbf{0.893} \\
\rowcolor{mygray}OmniAVS & \textbf{3.994} & \textbf{0.554} & \textbf{0.752} & \textbf{0.939} & {\ul 1.220} \\ \bottomrule
\end{tabular}
\end{adjustbox}
\vspace{-0.2cm}
\end{table}

\begin{table}[t!]
\setlength\tabcolsep{2pt} 
\setlength{\abovecaptionskip}{0cm}
\caption{Performance comparison of different saliency prediction models on the LJ-ODV database. ``FT" denotes fine-tuning on LJ-ODV based on pre-trained model weights. ``w/o p" and ``w/ p" indicate without and with pre-training on traditional audio-visual datasets, respectively. \label{tab:ljodv}}
\centering
\begin{adjustbox}{width=0.48\textwidth}
\begin{tabular}{l|cccccc|cccccc}
\toprule
Audio Mode & \multicolumn{6}{c|}{Stereo} & \multicolumn{6}{c}{Mute} \\ \midrule
Model\textbackslash{}Metric & {\scriptsize NSS↑} &{\scriptsize SIM↑} &{\scriptsize CC↑} &{\scriptsize AUC-J↑} &{\scriptsize s-AUC↑} &{\scriptsize KLD↓ }&{\scriptsize NSS↑} & {\scriptsize SIM↑}&{\scriptsize CC↑} &{\scriptsize AUC-J↑} &{\scriptsize s-AUC↑} &{\scriptsize KLD↓} \\ \midrule
IT \cite{730558}& 0.4776& 0.1312 & 0.0906 & 0.6301 & 0.5044 & 23.866 & 0.4495& 0.1505 & 0.1005 & 0.6292 & 0.5011 & 23.945 \\
GBVS \cite{10.5555/2976456.2976525}& 1.5647& 0.2215 & 0.2990 & 0.8652 & 0.5540 & 8.9961 & 1.3551& 0.2568 & 0.3136 & 0.8397 & 0.5021 & 9.1119 \\
SR \cite{SR}& 1.3868& 0.2175 & 0.2724 & 0.8539 & 0.5409 & 9.0187 & 1.2661& 0.2572 & 0.2958 & 0.8393 & 0.5074 & 9.0744 \\
SMVJ \cite{SMVJ}& 1.5884& {\ul 0.2255} & 0.3040 & 0.8710 & 0.5506 & 8.9694 & 1.3668& {\ul 0.2595} & 0.3158 & 0.8419 & 0.4916 & 9.0963 \\
SUN \cite{SUN}& {\ul 1.6778}& 0.2046 & 0.2934 & {\ul 0.8764} & \textbf{0.5814} & {\ul 8.9524} & \textbf{1.6300}& 0.2487 & 0.3321 & \textbf{0.8677} & {\color[HTML]{FF0000} \textbf{0.5646}} & \textbf{8.9758} \\
PQFT \cite{PQFT}& 1.3208& 0.2149 & 0.2524 & 0.8421 & 0.5269 & 9.0265 & 1.2896& 0.2594 & 0.2909 & 0.8366 & 0.5121 & 9.0427 \\
Judd \cite{Judd}& 1.6332& 0.1990 & {\ul 0.3097} & \textbf{0.8941} & {\ul 0.5727} & 9.0484 & {\ul 1.4574}& 0.2349 & {\ul 0.3325} & {\ul 0.8660} & 0.5029 & 9.1122 \\
SWD \cite{SWD}& \textbf{1.7301}& 0.2181 & \textbf{0.3264} & 0.8722 & 0.5659 & 8.9619 & 1.4567& 0.2498 & \textbf{0.3333} & 0.8402 & 0.4960 & 9.0870 \\
Murray \cite{Murray}& 1.3904& 0.1916 & 0.2691 & 0.8755 & 0.5689 & 9.1158 & 1.3136& 0.2312 & 0.3020 & 0.8578 & {\ul 0.5295} & 9.1473 \\
HFT \cite{HFT}& 1.5692& \textbf{0.2396} & 0.2960 & 0.8724 & 0.5534 & \textbf{8.9102} & 1.4034& \textbf{0.2784} & 0.3175 & 0.8521 & 0.5028 & {\ul 9.0103} \\ \midrule
TASED \cite{min2019tased}& {\ul 2.7332}& {\ul 0.3642} & {\ul 0.4663} & 0.8765 & \textbf{0.5878} & 10.981 & \textbf{2.0058} & \textbf{0.3459} & \textbf{0.4198} & {\ul 0.8235} & \textbf{0.5177} & {\ul 13.150} \\
STAViS \cite{tsiami2020stavis}& 2.5314& 0.3402 & 0.4506 & {\ul 0.8886} & {\ul 0.5831} & \textbf{9.3554} & {\ul 1.7433} & {\ul 0.3276} & {\ul 0.3910} & \textbf{0.8383} & {\ul 0.4959} & \textbf{10.113} \\
DAVE \cite{dave}& \textbf{2.7556}& \textbf{0.3704} & \textbf{0.4851} & \textbf{0.8931} & 0.5789 & {\ul 9.8295} & - & - & - & - & - & - \\ \midrule
TASED(FT) \cite{min2019tased}& \textbf{3.6683}& {\color[HTML]{0070C0}  \textbf{0.4322}} & {\color[HTML]{0070C0} \textbf{0.5870}} & \textbf{0.9147} & {\color[HTML]{FF0000} \textbf{0.6127}} & {\ul 9.5429} & \textbf{2.9639} & {\color[HTML]{0070C0} \textbf{0.4163}} & {\color[HTML]{0070C0} \textbf{0.5575}} & {\ul 0.8773} & \textbf{0.5454} & {\ul 11.098} \\
STAViS(FT) \cite{tsiami2020stavis}& 2.9802 & 0.3733 & 0.4914 & {\ul 0.9075} & 0.5865 & \textbf{8.8522} & {\ul 2.1092}& {\ul 0.3597} & {\ul 0.4615} & \textbf{0.8798} & {\ul 0.5113} & \textbf{8.8507} \\
DAVE(FT) \cite{dave}& {\ul 3.2321} & {\ul 0.4004} & {\ul 0.5322} & 0.8743 & {\ul 0.5873} & 11.883 & - & - & - & - & - & - \\ \midrule
\rowcolor{mygray}OmniAVS(w/o p) & {\color[HTML]{0070C0} {\ul 3.9020}} & {\ul 0.4184} & {\ul 0.5604} & {\color[HTML]{0070C0} {\ul 0.9284}} & {\ul 0.6069} & {\color[HTML]{0070C0} {\ul 8.5652}} & {\color[HTML]{0070C0} {\ul 3.0161}} & {\ul 0.3959} & {\ul 0.5405} & {\color[HTML]{0070C0} {\ul 0.9128}} & {\ul 0.5523} & {\color[HTML]{0070C0} {\ul 8.3564}} \\
\rowcolor{mygray}OmniAVS(w/ p) & {\color[HTML]{FF0000} \textbf{4.4199}} & {\color[HTML]{FF0000} \textbf{0.4439}} & {\color[HTML]{FF0000} \textbf{0.6158}} & {\color[HTML]{FF0000} \textbf{0.9321}} & {\color[HTML]{0070C0} {\ul 0.6117}} & {\color[HTML]{FF0000} \textbf{8.3425}} & {\color[HTML]{FF0000} \textbf{3.2733}} & {\color[HTML]{FF0000} \textbf{0.4213}} & {\color[HTML]{FF0000} \textbf{0.5662}} & {\color[HTML]{FF0000} \textbf{0.9155}} & {\color[HTML]{0070C0} \textbf{0.5527}} & {\color[HTML]{FF0000} {\ul \textbf{8.3472}}} \\ \bottomrule
\end{tabular}
\end{adjustbox}
\vspace{-0.2cm}
\end{table}

\vspace{-0.1cm}\subsection{Comparison with State-of-the-Arts on Visual-Only ODV Saliency Databases}
We evaluate the visual part of OmniAVS on two visual-only ODV saliency datasets, PVS-HM \cite{8418756} and Zhang-ODV \cite{Zhang_2018_ECCV}. We compare it against the visual components of other state-of-the-art audio-visual saliency prediction models (DAVE-V \cite{dave}, AVS360-V \cite{avs360}, SVGC-AVA-V \cite{10224292}) and visual-only saliency prediction models (TASED \cite{min2019tased}, SalGAN360 \cite{8551543}, offline-DHP \cite{8418756}, CP360 \cite{8578252}, SST-Sal \cite{BERNALBERDUN2022200}, ATSal \cite{10.1007/978-3-030-68796-0_22}, Spherical U-Net \cite{Zhang_2018_ECCV}). 
For both datasets, each model is trained on their respective training sets and tested on the test sets. 
\begin{table}[t!]
\vspace{-0.1cm}
\setlength{\abovecaptionskip}{0cm}
\setlength{\belowcaptionskip}{-0.2cm}
\caption{Performance comparison of different saliency prediction models on the PVS-HM database.  \label{tab:pvshm}}
\centering
\begin{adjustbox}{width=0.35\textwidth}
\begin{tabular}{l|ccccc}
\toprule
Model\textbackslash{}Metric & NSS↑ & SIM↑ & CC↑ & AUC-J↑ & KLD↓ \\ \midrule
TASED \cite{min2019tased}& 2.413 & - & 0.651 & 0.905 & - \\
SalGAN360 \cite{8551543}& 1.603 & - & 0.443 & - & - \\
offline-DHP \cite{8418756}& 3.275 & - & 0.704 & - & - \\
CP360 \cite{8578252}& 1.244 & 0.231 & 0.286 & - & 1.988 \\
SST-Sal \cite{BERNALBERDUN2022200}& 1.827 & - & 0.413 & - & - \\
ATSal \cite{10.1007/978-3-030-68796-0_22}& 2.489 & - & 0.564 & {\ul 0.914} & - \\
DAVE-V \cite{dave}& 3.176 & 0.519 & 0.690 & - & 1.038 \\
Spherical U-Net \cite{Zhang_2018_ECCV}& 3.289 & - & {\ul 0.767} & - & - \\
AVS360-V \cite{avs360}& 3.223 & 0.528 & 0.688 & - & 1.375 \\
SVGC-AVA-V \cite{10224292}& \textbf{3.554} & {\ul 0.547} & 0.731 & - & \textbf{0.738} \\
\rowcolor{mygray} OmniAVS & {\ul 3.487} & \textbf{0.621} & \textbf{0.795} & \textbf{0.922} & {\ul 0.974} \\ \bottomrule
\end{tabular}
\end{adjustbox}
\vspace{-0.2cm}
\end{table}

\begin{table}[t!]
\setlength{\abovecaptionskip}{0cm}
\setlength{\belowcaptionskip}{-0.2cm}
\setlength\tabcolsep{4pt} 
\caption{Performance comparison of different saliency prediction models on the Zhang-ODV database. \label{tab:zhangodv}}
\centering
\begin{adjustbox}{width=0.4\textwidth}
\begin{tabular}{l|cccccc}
\toprule
Model\textbackslash{}Metric & NSS↑ & SIM↑ & CC↑ & AUC-J↑ & s-AUC↑ & KLD↓ \\ \midrule
GBVS \cite{10.5555/2976456.2976525}& 0.8003 & - & 0.1254 & 0.7799 & - & - \\
LDS \cite{7407672}& 1.6589 & - & 0.2727 & 0.8169 & - & - \\
SalNet \cite{Pan_2016_CVPR}& 1.3958 & - & 0.2404 & 0.8266 & - & - \\
SALICON \cite{7410395}& 1.3178 & - & 0.2171 & 0.8074 & - & - \\
ACLNet \cite{aclnet}& 1.5869 & - & 0.2929 & 0.7906 & - & - \\
SaltiNet \cite{8265485}& 1.4470 & - & 0.2582 & 0.8579 & - & - \\
PanoSalNet \cite{panosalnet2018}& 2.9814 & - & 0.4892 & 0.6326 & - & - \\
Spherical U-Net \cite{Zhang_2018_ECCV}& {\ul 3.5340} & - & {\ul 0.6246} & {\ul 0.8977} & - & - \\
\rowcolor{mygray}OmniAVS & \textbf{3.9625} & \textbf{0.4694} & \textbf{0.6271} & \textbf{0.9348} & \textbf{0.6863} & \textbf{8.1119} \\ \bottomrule
\end{tabular}
\end{adjustbox}
\vspace{-0.2cm}
\end{table}

Tables \ref{tab:pvshm} and \ref{tab:zhangodv} demonstrate that the visual component of OmniAVS outperforms both the visual components of audio-visual saliency prediction models and the visual-only saliency prediction models, further confirming the robustness and effectiveness of the visual part of OmniAVS.

\subsection{Comparison with State-of-the-Arts on Traditional Audio-Visual Video Saliency Databases}
We also evaluate the performance of OmniAVS and  15 other state-of-the-art models on six traditional audio-visual saliency datasets: DIEM \cite{Mital2011ClusteringOG}, Coutrot1 \cite{Coutrot2014HowSF}, Coutrot2 \cite{Coutrot2016}, AVAD \cite{7457921}, SumMe \cite{Gygli2014CreatingSF}, and ETMD \cite{10.1016/j.image.2015.08.004}. The tested algorithms include 12 visual-only models (SALICON \cite{7410395}, SalNet \cite{Pan_2016_CVPR}, SalGAN \cite{Pan_2017_SalGAN}, DVA \cite{8240654}, SAM \cite{SAM}, Vavdvsm \cite{He2019UnderstandingAV}, DINet \cite{8868198}, DeepVS \cite{Jiang_2018_ECCV}, SalEMA \cite{salema}, TASED \cite{min2019tased}, TBA \cite{9157775}, ACLNet \cite{aclnet}) and 3 audio-visual models (STAViS \cite{tsiami2020stavis}, DAVF \cite{9506089}, LAVS \cite{10.1145/3576857}). For experiments with OmniAVS, we remove the orientation injection part of the audio-visual feature module. The results, as shown in Table \ref{tab:trad}, demonstrate the strong generality of our OmniAVS net and confirm its ability to predict saliency in traditional audio-visual videos.
\begin{table*}[t!]
\setlength{\abovecaptionskip}{0cm}
\setlength{\belowcaptionskip}{-0.2cm}
\setlength\tabcolsep{5pt} 
\caption{Performance comparison of different saliency prediction models on 6 traditional audio-visual video saliency databases: DIEM \cite{Mital2011ClusteringOG}, Coutrot1 \cite{Coutrot2014HowSF}, Coutrot2 \cite{Coutrot2016}, AVAD \cite{7457921}, SumMe \cite{Gygli2014CreatingSF}, and ETMD \cite{10.1016/j.image.2015.08.004}. \label{tab:trad}}
\centering
\resizebox{0.8\linewidth}{!}{
\begin{tabular}{l|ccccc|ccccc|ccccc}
\toprule
Database & \multicolumn{5}{c|}{DIEM \cite{Mital2011ClusteringOG}} & \multicolumn{5}{c|}{Coutrot1 \cite{Coutrot2014HowSF}} & \multicolumn{5}{c}{Coutrot2 \cite{Coutrot2016}} \\ \midrule
Model\textbackslash{}Metric & {\scriptsize NSS↑} &{\scriptsize SIM↑} &{\scriptsize CC↑} & {\scriptsize AUC-J↑} & {\scriptsize s-AUC↑}& {\scriptsize NSS↑} &{\scriptsize SIM↑} &{\scriptsize CC↑ }&{\scriptsize AUC-J↑} & {\scriptsize s-AUC↑} &{\scriptsize NSS↑} & {\scriptsize SIM↑} &{\scriptsize CC↑} & {\scriptsize AUC-J↑} & {\scriptsize s-AUC↑} \\ \midrule
SALICON \cite{7410395}& 1.96 & 0.3821 & 0.4782 & 0.8552 & 0.6410 & 1.92 & 0.3428 & 0.4316 & 0.8547 & 0.5761 & 3.05 & 0.2932 & 0.4631 & 0.9321 & 0.6322 \\
SalNet \cite{Pan_2016_CVPR} & 1.52 & 0.3183 & 0.4075 & 0.8321 & 0.6227 & 1.41 & 0.2732 & 0.3402 & 0.8248 & 0.5597 & 1.82 & 0.2019 & 0.3012 & 0.8966 & 0.6000 \\
SalGAN \cite{Pan_2017_SalGAN} & 1.89 & 0.3931 & 0.4868 & 0.8570 & 0.6609 & 1.85 & 0.3321 & 0.4161 & 0.8536 & 0.5799 & 2.96 & 0.2909 & 0.4398 & 0.9331 & 0.6183 \\
DVA \cite{8240654} & 1.97 & 0.3785 & 0.4779 & 0.8547 & 0.6410 & 2.07 & 0.3324 & 0.4306 & 0.8531 & 0.5783 & 3.45 & 0.2742 & 0.4634 & 0.9328 & 0.6324 \\
SAM \cite{SAM} & 2.05 & 0.4261 & 0.4930 & 0.8592 & 0.6446 & 2.11 & 0.3672 & 0.4329 & 0.8571 & 0.5768 & 3.02 & 0.3041 & 0.4194 & 0.9320 & 0.6152 \\
Vavdvsm \cite{He2019UnderstandingAV} & 2.12 & 0.4052 & 0.5031 & 0.8712 & 0.6485 & 2.07 & 0.3686 & 0.4328 & 0.8569 & 0.5776 & 3.06 & 0.3684 & 0.4325 & 0.8562 & 0.5776 \\
DINet \cite{8868198} & 2.15 & 0.4183 & 0.5125 & 0.8820 & 0.6581 & 2.09 & 0.3691 & 0.4331 & 0.8572 & 0.5783 & 3.16 & 0.3687 & 0.4328 & 0.8569 & 0.5780 \\
DeepVS \cite{Jiang_2018_ECCV} & 1.86 & 0.3923 &  0.4523& 0.8406 & 0.6256 & 1.77 & 0.3174 & 0.3595 & 0.8306 & 0.5617 & 3.79 & 0.2590 & 0.4494 & 0.9255 & 0.6469 \\
SalEMA \cite{salema}& 2.01 & 0.4018 & 0.5012 & 0.8695 & 0.6466 & 2.04 & 0.3678 & 0.4319 & 0.8553 & 0.5762 & 3.02 & 0.3675 & 0.4322 & 0.8558 & 0.5762 \\
TASED \cite{min2019tased}& 2.16 & 0.4615 & 0.5579 & 0.8812 & 0.6579 & 2.18 & 0.3884 & 0.4799 & 0.8676 & 0.5808 & 3.17 & 0.3142 & 0.4375 & 0.9216 & 0.6118 \\
TBA \cite{9157775} & 2.04 & 0.4025 & 0.5022 & 0.8699 & 0.6472 & 1.97 & 0.3683 & 0.4324 & 0.8564 & 0.5754 & 2.99 & 0.3663 & 0.4319 & 0.8558 & 0.5768 \\
ACLNet \cite{aclnet} & 2.02 & 0.4279 & 0.5229 & 0.8690 & 0.6221 & 1.92 & 0.3612 & 0.4253 & 0.8502 & 0.5429 & 3.16 & 0.3229 & 0.4485 & 0.9267 & 0.5943 \\
STAViS \cite{tsiami2020stavis}& 2.26 & 0.4824 & 0.5795 & 0.8838 & 0.6741 & 2.11 & 0.3935 & 0.4722 & 0.8686 & 0.5847 & 5.28 & 0.5111 & {\ul 0.7349} & 0.9581 & 0.7106 \\
DAVF \cite{9506089} & {\ul 2.37} & {\ul 0.5075} & {\ul 0.6120} & \textbf{0.8954} & {\ul 0.6829} & {\ul 2.22} & {\ul 0.4167} & {\ul 0.5081} & {\ul 0.8704} & 0.5810 & \textbf{5.87} & \textbf{0.5715} & \textbf{0.7789} & \textbf{0.9612} & \textbf{0.7186} \\
LAVS \cite{10.1145/3576857} & 2.27 & 0.4823 & 0.5796 & 0.8840 & 0.6744 & 2.19 & 0.3938 & 0.4806 & 0.8689 & {\ul 0.5885} & 4.98 & {\ul 0.5115} & 0.7336 & 0.9571 & {\ul 0.7109} \\
\rowcolor{mygray}OmniAVS & \textbf{2.51} & \textbf{0.5253} & \textbf{0.6207} & {\ul 0.8919} & \textbf{0.6868} & \textbf{2.54} & \textbf{0.4318} & \textbf{0.5229} & \textbf{0.8762} & \textbf{0.5939} & {\ul 5.29} & 0.4681 & 0.6538 & {\ul 0.9590} & 0.6967 \\ \midrule
Database & \multicolumn{5}{c|}{AVAD \cite{7457921}} & \multicolumn{5}{c|}{SumMe \cite{Gygli2014CreatingSF}} & \multicolumn{5}{c}{ETMD \cite{10.1016/j.image.2015.08.004}} \\ \midrule
Model\textbackslash{}Metric & {\scriptsize NSS↑} &{\scriptsize SIM↑} &{\scriptsize CC↑} & {\scriptsize AUC-J↑} & {\scriptsize s-AUC↑}& {\scriptsize NSS↑} &{\scriptsize SIM↑} &{\scriptsize CC↑ }&{\scriptsize AUC-J↑} & {\scriptsize s-AUC↑} &{\scriptsize NSS↑} & {\scriptsize SIM↑} &{\scriptsize CC↑} & {\scriptsize AUC-J↑} & {\scriptsize s-AUC↑} \\ \midrule
SALICON \cite{7410395} & 2.94 & 0.3712 & 0.5246 & 0.8885 & 0.5826 & 2.16 & 0.2922 & 0.3991 & 0.8681 & 0.6692 & 2.68 & 0.3272 & 0.4974 & 0.9041 & 0.7292 \\
SalNet \cite{Pan_2016_CVPR} & 1.85 & 0.2564 & 0.3831 & 0.8690 & 0.5616 & 1.55 & 0.2274 & 0.3320 & 0.8488 & 0.6451 & 1.90 & 0.2253 & 0.3879 & 0.8897 & 0.6992 \\
SalGAN \cite{Pan_2017_SalGAN} & 2.55 & 0.3608 & 0.4912 & 0.8865 & 0.5799 & 1.97 & 0.2897 & 0.3978 & 0.8754 & {\ul 0.6882} & 2.46 & 0.3117 & 0.4765 & 0.9035 & {\ul 0.7463} \\
DVA \cite{8240654} & 3.00 & 0.3633 & 0.5247 & 0.8887 & 0.5820 & 2.14 & 0.2811 & 0.3983 & 0.8681 & 0.6686 & 2.72 & 0.3165 & 0.4965 & 0.9039 & 0.7288 \\
SAM \cite{SAM}& 2.99 & 0.4244 & 0.5279 & 0.9025 & 0.5777 & 2.21 & 0.3272 & 0.4041 & 0.8717 & 0.6728 & 2.78 & 0.3790 & 0.5068 & 0.9073 & 0.7310 \\
Vavdvsm \cite{He2019UnderstandingAV} & 2.95 & 0.4237 & 0.5243 & 0.8882 & 0.5819 & 2.23 & 0.3312 & 0.4052 & 0.8781 & 0.6782 & 2.66 & 0.3543 & 0.4791 & 0.9052 & 0.7324 \\
DINet \cite{8868198} & 2.99 & 0.4241 & 0.5247 & 0.8886 & 0.5824 & 2.25 & {\ul 0.3521} & 0.4123 & 0.8812 & 0.6785 & 2.69 & 0.3792 & 0.5041 & 0.9065 & 0.7332 \\
DeepVS \cite{Jiang_2018_ECCV}& 3.01 & 0.3914 & 0.5281 & 0.8968 & 0.5858 & 1.62 & 0.2622 & 0.3172 & 0.8422 & 0.6120 & 2.48 & 0.3495 & 0.4616 & 0.9041 & 0.6861 \\
SalEMA \cite{salema}& 2.91 & 0.4229 & 0.5240 & 0.8875 & 0.5812 & 2.18 & 0.3287 & 0.4018 & 0.8764 & 0.6766 & 2.54 & 0.3512 & 0.4781 & 0.9048 & 0.7321 \\
TASED \cite{min2019tased}& 3.16 & 0.4395 & 0.6006 & 0.9146 & 0.5898 & 2.10 & 0.3337 & 0.4288 & 0.8840 & 0.6570 & 2.63 & 0.3660 & 0.5093 & 0.9164 & 0.7117 \\
TBA \cite{9157775} & 2.86 & 0.4234 & 0.5198 & 0.8876 & 0.5775 & 2.21 & 0.3328 & 0.4036 & 0.8792 & {\ul 0.6882} & 2.56 & 0.3628 & 0.4642 & 0.9038 & 0.7316 \\
ACLNet \cite{aclnet}& 3.17 & 0.4463 & 0.5809 & 0.9053 & 0.5600 & 1.79 & 0.2965 & 0.3795 & 0.8687 & 0.6092 & 2.36 & 0.3290 & 0.4771 & 0.9152 & 0.6752 \\
STAViS \cite{tsiami2020stavis}& 3.18 & 0.4578 & 0.6086 & 0.9196 & 0.5936 & 2.04 & 0.3373 & 0.4220 & 0.8883 & 0.6562 & {\ul 2.94} & 0.4251 & 0.5690 & 0.9316 & 0.7317 \\
DAVF \cite{9506089} & {\ul 3.31} & {\ul 0.4816} & {\ul 0.6212} & \textbf{0.9232} & 0.5927 & 2.15 & 0.3496 & {\ul 0.4451} & {\ul 0.8951} & 0.6684 & 2.91 & 0.4147 & 0.5991 & 0.9314 & 0.7319 \\
LAVS \cite{10.1145/3576857} & 3.19 & 0.4576 & 0.6073 & 0.9186 & {\ul 0.5938} & {\ul 2.28} & 0.3381 & 0.4225 & 0.8886 & \textbf{0.6927} & {\ul 2.94} & {\ul 0.4252} & {\ul 0.5997} & {\ul 0.9320} & 0.7401 \\
\rowcolor{mygray}OmniAVS & \textbf{3.71} & \textbf{0.5194} & \textbf{0.6710} & {\ul 0.9231} & \textbf{0.6014} & \textbf{2.54} & \textbf{0.3866} & \textbf{0.4936} & \textbf{0.9065} & 0.6856 & \textbf{3.31} & \textbf{0.4728} & \textbf{0.6141} & \textbf{0.9366} & \textbf{0.7539} \\ \bottomrule
\end{tabular}
}
\vspace{-0.3cm}
\end{table*}

\subsection{Ablation Study}
We conduct ablation experiments to assess the effectiveness of the architecture, feature fusion module, and the temporal module of OmniAVS. All experiments for this part are performed under the ambisonics mode of the AVS-ODV database.

\subsubsection{Architecture}
We conduct ablation studies focusing on the architecture from four perspectives: the visual backbone, the hierarchical structure, the depth of the U-net architecture, and the method of frame sampling from the input video sequences. The results are shown in Table \ref{tab:ve}.
A decrease in performance metrics is observed after replacing the visual backbone with CLIP (ViT-L/14) \cite{CLIP}, which indicates that the multimodal joint embedding space provided by ImageBind \cite{girdhar2023imagebind} effectively enhances the model's ability to utilize the features of the auditory modality. We simplify the hierarchical structure of the network by using only single-scale visual features from the final output of the visual backbone and include only one audio-visual saliency estimation block, and see a significant decrease in model performance.
Inspired by Li \textit{et al.} \cite{Li2022ExploringPV}, we replace the encoder part of the U-Net architecture with a plain backbone that employs a simple feature pyramid without an FPN, which transforms features only from the final output of the visual backbone to obtain multi-scale features. However, this approach proves ineffective for our task, possibly due to the high dependency of saliency prediction on lower-level image features. 

Moreover, we conduct ablation studies on the depth of the U-Net structure. Our model (``Base" version) employs visual features from seven stages, comprising six levels of skip connections. In the ``Small" model variant, the architecture is reduced to include visual features from four stages, with three levels of skip connections. Conversely, in the ``Large" variant, the number of stages is increased to eight, with seven levels of skip connections. The ``Small" version exhibits a discernible decline in performance. The ``Large" version does not demonstrate a commensurate increase in performance with the increase in model parameters, suggesting overfitting. These findings validate our choice of a U-Net with six skip connection levels as the optimal configuration.
Additionally, we conduct ablation experiments on the frame sampling method for input videos. The frame interval, denoted as ``Step", determines the temporal span between adjacent input frames, while ``Num" refers to the total number of input video frames, which directly influences the computational load of the visual encoder. 
Experimental results indicate that the model performs optimally when Step is 8 and Num is also 8, balancing performance with computational efficiency.


\begin{table}[t!]
\setlength{\abovecaptionskip}{0cm}
\setlength{\belowcaptionskip}{-0.2cm}
\setlength\tabcolsep{5pt} 
\caption{The ablation study of the architecture.\label{tab:ve}}
\centering
\begin{adjustbox}{width=0.45\textwidth}
\begin{tabular}{c|l|ccccc}
\toprule
\multicolumn{1}{l|}{} & Method & NSS↑ & SIM↑ & CC↑ & AUC-J↑ & s-AUC↑ \\ \midrule
\multicolumn{1}{c|}{Visual Backbone} & CLIP & 4.9144 & 0.4739 & 0.6321 & 0.9542 & 0.6293 \\ \midrule
\multirow{2}{*}{Hierarchical Structure} & Simple & 4.2299 & 0.4533 & 0.6031 & 0.9505 & 0.6105 \\
& Plain & 4.6061 & 0.4703 & 0.6169 & 0.9526 & 0.6184 \\ \midrule
\multirow{2}{*}{U-Net Depth} & Small & 4.9111 & 0.4781 & 0.6320 & 0.9540 & 0.6339 \\
 & Large & 4.9665 & 0.4789 & 0.6393 & 0.9531 & 0.6327 \\ \midrule
\multirow{8}{*}{Frame Sampling} & Step=1,Num=4 & 4.8867 & 0.4701 & 0.6289 & 0.9546 & 0.6286 \\
 & Step=1,Num=8 & 4.9461 & 0.4720 & 0.6263 & 0.9525 & 0.6275 \\
 & Step=1,Num=16 & 4.9758 & 0.4751 & 0.6342 & {\ul0.9548} & 0.6355 \\
 & Step=4,Num=4 & 4.9778 & 0.4768 & 0.6347 & 0.9538 & 0.6339 \\
 & Step=4,Num=8 & 5.0276 & 0.4779 & 0.6377 & 0.9538 & 0.6350 \\
 & Step=4,Num=16 & 5.0354 & 0.4772 & 0.6395 & 0.9545 & {\ul 0.6393} \\
 & Step=8,Num=4 & {\ul 5.0394} & {\ul 0.4807} & 0.6402 & 0.9539 & 0.6373 \\
 & Step=8,Num=16 & 5.0263 & 0.4771 & {\ul 0.6410} & \textbf{0.9551} & \textbf{0.6415} \\ \midrule
  \rowcolor{mygray}
\multicolumn{2}{c|}{ImageBind+U-Net(Base); Step=8,Num=8}   & \textbf{5.0703} & \textbf{0.4838} & \textbf{0.6452} & 0.9541 & 0.6357 \\ \bottomrule
\end{tabular}\vspace{-0.15cm}
\end{adjustbox}
\vspace{-0.3cm}
\end{table}

\subsubsection{Temporal Module}
To evaluate the effectiveness of the temporal module, ablation studies are conducted focusing on two aspects, \textit{i.e.,} the choice of the temporal module and the number of levels at which it operates. Initially, the model is tested without the temporal module. Subsequently, the spatio-temporal Transformer is replaced with the other two temporal aggregation methods described in Section \ref{sec:temporal}. Additionally, the number of levels in the visual backbone at which the temporal module operates is varied. Specifically, the model's performance is tested when the temporal module is applied at one level (\textit{i.e.}, after the $i$-th stage where $i\in\{7\}$), two levels (\(i\in\{6, 7\}\)), and four levels (\(i\in\{4, 5, 6, 7\}\)). According to the results in Table \ref{tab:temporal}, a noticeable decline in performance across all metrics is observed when the temporal module is removed. Although
applying 3D ConvNet or spatio-temporal GRU results in slight improvements compared to without temporal integration, these methods do not surpass the performance achieved with the spatio-temporal Transformer. Furthermore, the model performs optimally when the temporal module is applied at three levels. Consequently, we apply the temporal module at the last three stages (\(i \in \{5, 6, 7\}\)).

\begin{table}[t!]
\setlength{\abovecaptionskip}{0cm}
\setlength{\belowcaptionskip}{-0.2cm}
\setlength\tabcolsep{5pt} 
\caption{The ablation study of the temporal module.\label{tab:temporal}}
\centering
\begin{adjustbox}{width=0.46\textwidth}
\begin{tabular}{c|l|ccccc}
\toprule
\multicolumn{1}{l|}{} & Method & NSS↑ & SIM↑ & CC↑ & AUC-J↑ & s-AUC↑ \\ \midrule
\multirow{3}{*}{Temporal Module} & None & 4.8780 & 0.4675 & 0.6241 & 0.9530 & 0.6264 \\
 & 3D ConvNet & 4.9483 & 0.4807 & 0.6364 & 0.9534 & 0.6347 \\
 & Spatio-Temporal GRU & {\ul5.0155} & 0.4813 & 0.6393 & 0.9534 & {\ul0.6369} \\ \midrule
\multirow{3}{*}{Temporal Level} & 1 & 4.9332 & 0.4770 & 0.6334 & 0.9528 & 0.6353 \\
 & 2 &  5.0010 & {\ul 0.4821} & {\ul 0.6403} & 0.9534 & \textbf{0.6390} \\
 & 4 & 4.9556 & 0.4714 & 0.6305 & {\ul0.9537} & 0.6284 \\ \midrule
 \rowcolor{mygray}
\multicolumn{2}{c|}{Spatio-Temporal Transformer$\times$3} & \textbf{5.0703} & \textbf{0.4838} & \textbf{0.6452} & \textbf{0.9541} & 0.6357 \\ \bottomrule
\end{tabular}
\end{adjustbox}
\vspace{-0.2cm}
\end{table}

\subsubsection{Fusion Module}
\begin{figure}[t!]
\centering
\subfloat[multi-level audio features]{\includegraphics[width=0.49\linewidth]{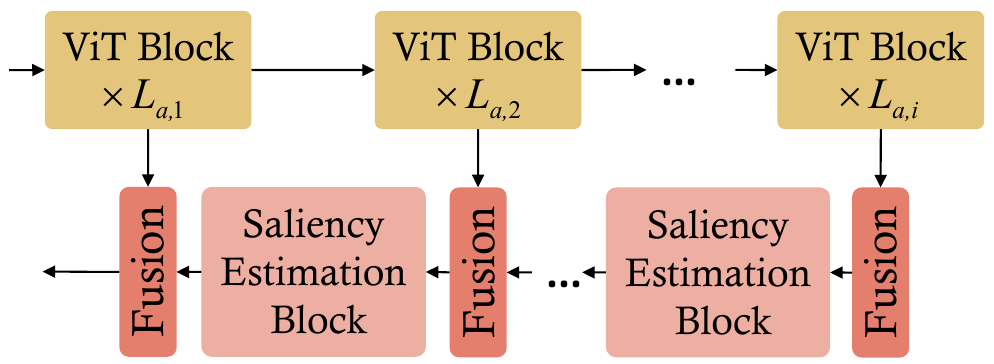}} \hspace{1pt}
\subfloat[single-level audio features]{\includegraphics[width=0.49\linewidth]{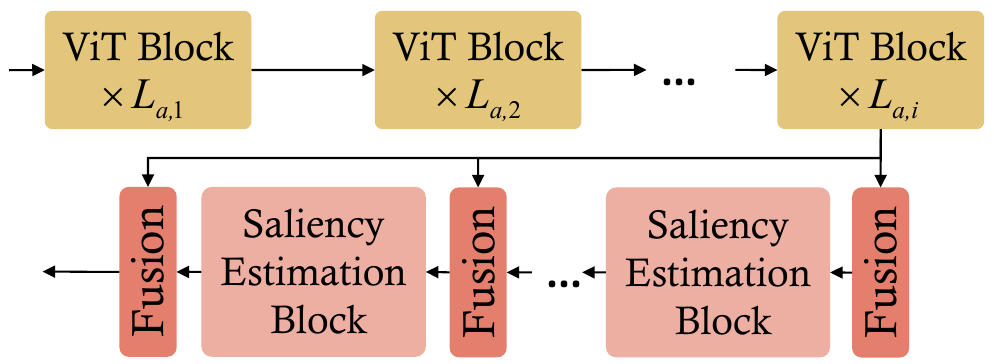}}
\caption{The injection of multi-level and single-level audio features.}
\label{fig:multisingle}
\vspace{-0.2cm}
\end{figure}

To evaluate the rationality of the feature fusion module, we conduct ablation studies on both the fusion approach and the fusion level. First, the fusion module is removed entirely, resulting in a significant decrease in model performance, as shown in Table \ref{tab:fusion}. This observation indicates that audio information is essential for predicting audio-visual saliency in ODVs, and our model effectively integrates features from both modalities. 
Next, we explore alternative fusion approaches. Replacing the fusion method with simple concatenation and channel-adjusting convolution caused a significant drop in performance across all metrics, indicating that simple concatenation fails to fully capture the complex relationships between audio and visual modalities. Removing the directional audio information and the corresponding orientation injection leads to a decline in performance, suggesting that these cues contribute to the model’s effectiveness. In contrast, using only directional audio information by replacing the features extracted from the audio encoder with AEM and injecting it through the CrossTransformer results in a more significant performance drop, highlighting the greater importance of semantic audio features. Nevertheless, in these two cases, the performance still surpasses that of the unimodal results, demonstrating that both semantic and directional audio information are crucial. Our model effectively integrates both, leading to enhanced overall performance. 
Finally, the number of levels at which the fusion module operates is also varied, and the effects of injecting multi-level versus single-level audio features are compared, as illustrated in Fig. \ref{fig:multisingle}. As the number of levels increases, the model's performance is enhanced, and multi-level audio feature injection surpasses that of single-level. Given the marginal performance gains from integrating audio features at four levels compared to three, and considering computational efficiency, we finally inject multi-level audio features through the CrossTransformer at three levels.

\begin{table}[t!]
\setlength{\abovecaptionskip}{0cm}
\setlength{\belowcaptionskip}{-0.2cm}
\setlength\tabcolsep{4.5pt} 
\caption{The ablation study of the fusion module. \label{tab:fusion}}
\centering
\begin{adjustbox}{width=0.46\textwidth}
\begin{tabular}{c|cccccc}
\toprule
\multicolumn{1}{l|}{} & \multicolumn{1}{l|}{Method} & NSS↑ & SIM↑ & CC↑ & AUC-J↑ & s-AUC↑ \\ \midrule
\multicolumn{1}{c|}{Modality} & \multicolumn{1}{l|}{Visual-Only} & 4.7599 & 0.4648 & 0.6125 & 0.9523 & 0.6220 \\ \midrule
\multirow{3}{*}{Fusion Approach} & \multicolumn{1}{l|}{Concat} & 4.8670 & 0.4738 & 0.6290 & 0.9530 & 0.6297 \\
& \multicolumn{1}{l|}{{No Orientation}}&{4.9356}&{0.4827}&{0.6358}&{0.9529} &{0.6362} \\
& \multicolumn{1}{l|}{{AEM}} & {4.9023}&{0.4641}&{0.6272}&{0.9544}&{0.6268}
 \\\midrule
\multirow{6}{*}{Fusion Level} & \multicolumn{1}{l|}{1} & 4.8592 & 0.4665 & 0.6201 & 0.9534 & 0.6247 \\
 & \multicolumn{1}{l|}{2 (single)} & 4.9055 & 0.4689 & 0.6257 & 0.9538 & 0.6264 \\
 & \multicolumn{1}{l|}{2 (multi)} & 4.9470 & 0.4701 & 0.6300 & 0.9538 & 0.6272 \\
 & \multicolumn{1}{l|}{3 (single)} & 5.0300 & \textbf{0.4862} & 0.6422 & 0.9537 & 0.6395 \\
 & \multicolumn{1}{l|}{4 (single)} & 5.0017 & {\ul 0.4861} & 0.6437 & {\ul 0.9549} & {\ul 0.6421} \\
 & \multicolumn{1}{l|}{4 (multi)} & {\ul 5.0507} & 0.4859& {\ul 0.6448} & \textbf{0.9549} & \textbf{0.6424} \\ \midrule
 \rowcolor{mygray}
\multicolumn{2}{c|}{CrossTransformer$\times$3 (multi)} & \textbf{5.0703} & 0.4838 & \textbf{0.6452 }& 0.9541 & 0.6357 \\ \bottomrule
\end{tabular}
\end{adjustbox}
\vspace{-0.1cm}
\end{table}

\subsubsection{Saliency Loss}
To assess the impact of different components of the saliency loss on model performance, we conduct an ablation study by experimenting with various combinations of loss function terms, while maintaining their empirically determined weights. The results in Table \ref{tab:loss} demonstrate that the model achieves optimal performance when all four loss components (L1, CC, KL, and BCE) are jointly utilized. The second-best performance is observed when only the L1 loss is removed, suggesting that while L1 loss is not the dominant contributor, it plays a complementary role in guiding pixel-level accuracy. The performance gradually degrades as more loss components are removed, with the most significant decline occurring when only the L1 loss is used. In this case, the model fails to learn meaningful saliency patterns, indicating that L1 loss alone is insufficient for the task. These results highlight the importance of a well-designed loss function in saliency prediction. While L1 loss provides a basic measure of pixel-wise error, it fails to capture higher-level statistical and structural properties of saliency maps. The CC loss ensures alignment with the ground truth distribution, the KL divergence refines probabilistic alignment, and the BCE loss further enhances the model's ability to distinguish salient from non-salient regions. Combining these four components enables our model to leverage their strengths, generating saliency maps that are both precise and perceptually plausible, leading to superior performance.

\begin{table}[t!]
\setlength{\abovecaptionskip}{0cm}
\setlength{\belowcaptionskip}{-0.2cm}
\setlength\tabcolsep{4.5pt} 
{\caption{The ablation study of the loss function.
\label{tab:loss}}}
\centering
\begin{adjustbox}{width=0.43\textwidth}
\begin{tabular}{cccc|ccccc}
\toprule
{ $\mathcal{L}_{\text{L1}}$} & { $\mathcal{L}_{\text{CC}}$} & { $\mathcal{L}_{\text{KL}}$} & { $\mathcal{L}_{\text{BCE}}$} & { NSS↑} & { SIM↑} & { CC↑} & { AUC-J↑} & { s-AUC↑} \\ \midrule
{ \checkmark} & { } & { } & { } & { 1.0149} & { 0.1523} & { 0.1897} & { 0.8080} & { 0.5105} \\
{ } & { \checkmark} & { } & { } & { \textbf{5.0983}} & { 0.4534} & { 0.6402} & { 0.9524} & { 0.6304} \\
{ } & { } & { \checkmark} & { } & { 4.8542} & { 0.4723} & { 0.6277} & { 0.9530} & { 0.6266} \\
{ } & { } & { } & { \checkmark} & { 4.2610} & { 0.4110} & { 0.5551} & { 0.9472} & { 0.6108} \\
{ \checkmark} & { \checkmark} & { } & { } & { 4.6967} & { 0.4478} & { 0.5881} & { 0.9421} & { 0.6178} \\
{ \checkmark} & { } & { \checkmark} & { } & { 4.8953} & { 0.4704} & { 0.6299} & { 0.9541} & { 0.6276} \\
{ \checkmark} & { } & { } & { \checkmark} & { 4.3844} & { 0.4228} & { 0.5561} & { 0.9452} & { 0.6085} \\
{ } & { \checkmark} & { \checkmark} & { } & { 4.9739} & { 0.4753} & { 0.6348} & { {\ul 0.9545}} & { 0.6304} \\
{ } & { \checkmark} & { } & { \checkmark} & { 4.8905} & { 0.4498} & { 0.6157} & { 0.9518} & { 0.6258} \\
{ } & { } & { \checkmark} & { \checkmark} & { 4.8605} & { 0.4748} & { 0.6322} & { 0.9540} & { 0.6264} \\
{ \checkmark} & { \checkmark} & { \checkmark} & { } & { 4.9939} & { 0.4783} & { 0.6354} & { 0.9536} & { 0.6286} \\
{ \checkmark} & { \checkmark} & { } & { \checkmark} & { 4.9404} & { 0.4581} & { 0.6162} & { 0.9512} & { 0.6250} \\
{ \checkmark} & { } & { \checkmark} & { \checkmark} & { 4.9058} & { 0.4749} & { 0.6290} & { 0.9529} & { 0.6324} \\
{ } & { \checkmark} & { \checkmark} & { \checkmark} & { 4.9607} & { {\ul 0.4811}} & { {\ul 0.6413}} & { \textbf{0.9546}} & { {\ul 0.6347}} \\
{ \checkmark} & { \checkmark} & { \checkmark} & { \checkmark} & { {\ul 5.0703}} & { \textbf{0.4838}} & { \textbf{0.6452}} & { 0.9541} & { \textbf{0.6357}} \\ \bottomrule
\end{tabular}
\end{adjustbox}
\vspace{-0.1cm}
\end{table}

\subsection{Cross-Database Evaluation}
To verify the generalization ability and robustness of our proposed OmniAVS model across different audio-visual saliency prediction scenarios, we conduct the cross-database evaluation between two datasets: AVS-ODV and YQ-ODV \cite{10224292}. The models are trained on the training set of one dataset and tested on the testing set of the other. 

The results in Table \ref{tab:cross} demonstrate the superior performance of OmniAVS compared to other state-of-the-art models. Specifically, when trained on AVS-ODV and tested on YQ-ODV \cite{10224292}, OmniAVS achieves the highest scores in most metrics. Similarly, when trained on YQ-ODV \cite{10224292} and tested on AVS-ODV, OmniAVS still exhibits significant enhancements across all metrics, confirming its consistency and reliability across datasets. Furthermore, when OmniAVS is trained only on AVS-ODV, its prediction performance on YQ-ODV \cite{10224292} is close to that achieved after training on the YQ-ODV \cite{10224292} training set. This suggests that AVS-ODV contains a variety of audio-visual scenarios that are sufficient to train models with high generalization ability. Conversely, the predictive performance of OmniAVS on AVS-ODV, when trained only on YQ-ODV \cite{10224292}, is significantly worse than the optimal performance shown in Table \ref{tab:avsodv}. This discrepancy can be attributed to the limited data volume and lack of diversity in YQ-ODV \cite{10224292}, which pose challenges for models trained solely on this database to effectively handle the more complex AVS-ODV database.

\begin{table}[t!]
\vspace{-0.2cm}
\setlength{\abovecaptionskip}{0cm} 
\setlength{\belowcaptionskip}{-0.2cm}
\setlength\tabcolsep{2pt} 
\caption{Cross-database evaluation performance comparisons on AVS-ODV and YQ-ODV.\label{tab:cross}}
\centering\arrayrulecolor{black}
\begin{adjustbox}{width=0.48\textwidth}
\begin{tabular}{l|cccccc|cccccc}
\toprule
Training & \multicolumn{6}{c|}{AVS-ODV} & \multicolumn{6}{c}{YQ-ODV} \\ \midrule
Testing & \multicolumn{6}{c|}{YQ-ODV} & \multicolumn{6}{c}{AVS-ODV} \\ \midrule
Model\textbackslash{}Metric & NSS↑ & SIM↑ & CC↑ & AUC-J↑ & s-AUC↑ & KLD↓ & NSS↑ & SIM↑ & CC↑ & AUC-J↑ & s-AUC↑ & KLD↓ \\ \midrule
TASED \cite{min2019tased}& {\ul 3.1557} & \textbf{0.4788} & {\ul 0.6139} & 0.9132 & {\ul 0.6451} & 8.3138 & {\ul 2.6711} & {\ul 0.3393} & {\ul 0.4216} & 0.8996 & {\ul 0.5645} & 9.4468 \\
SalEMA \cite{salema}& 3.1214 & 0.4713 & 0.6060 & \textbf{0.9270} & 0.6447 & \textbf{6.8347} & 2.4497 & 0.3186 & 0.3832 & {\ul 0.9116} & 0.5560 & {\ul 8.2491} \\
STAViS \cite{tsiami2020stavis}& 2.8053 & 0.4575 & 0.5809 & 0.9193 & 0.6422 & 7.4192 & 2.3392 & 0.3174 & 0.3820 & 0.9049 & 0.5559 & 8.4971 \\
 \rowcolor{mygray}
OmniAVS & \textbf{3.3977} & {\ul 0.4762} & \textbf{0.6250} & {\ul 0.9260} & \textbf{0.6495} & {\ul 7.0427} & \textbf{3.0528} & \textbf{0.3484} & \textbf{0.4617} & \textbf{0.9371} & \textbf{0.5868} & \textbf{7.8908} \\ \bottomrule
\end{tabular}
\end{adjustbox}
\vspace{-0.1cm}
\end{table}

\subsection{Computational Efficiency}
\begin{figure}[t!]
\setlength{\abovecaptionskip}{0cm}
\setlength{\belowcaptionskip}{-0.2cm}
\centering
\includegraphics[width=2.2in]{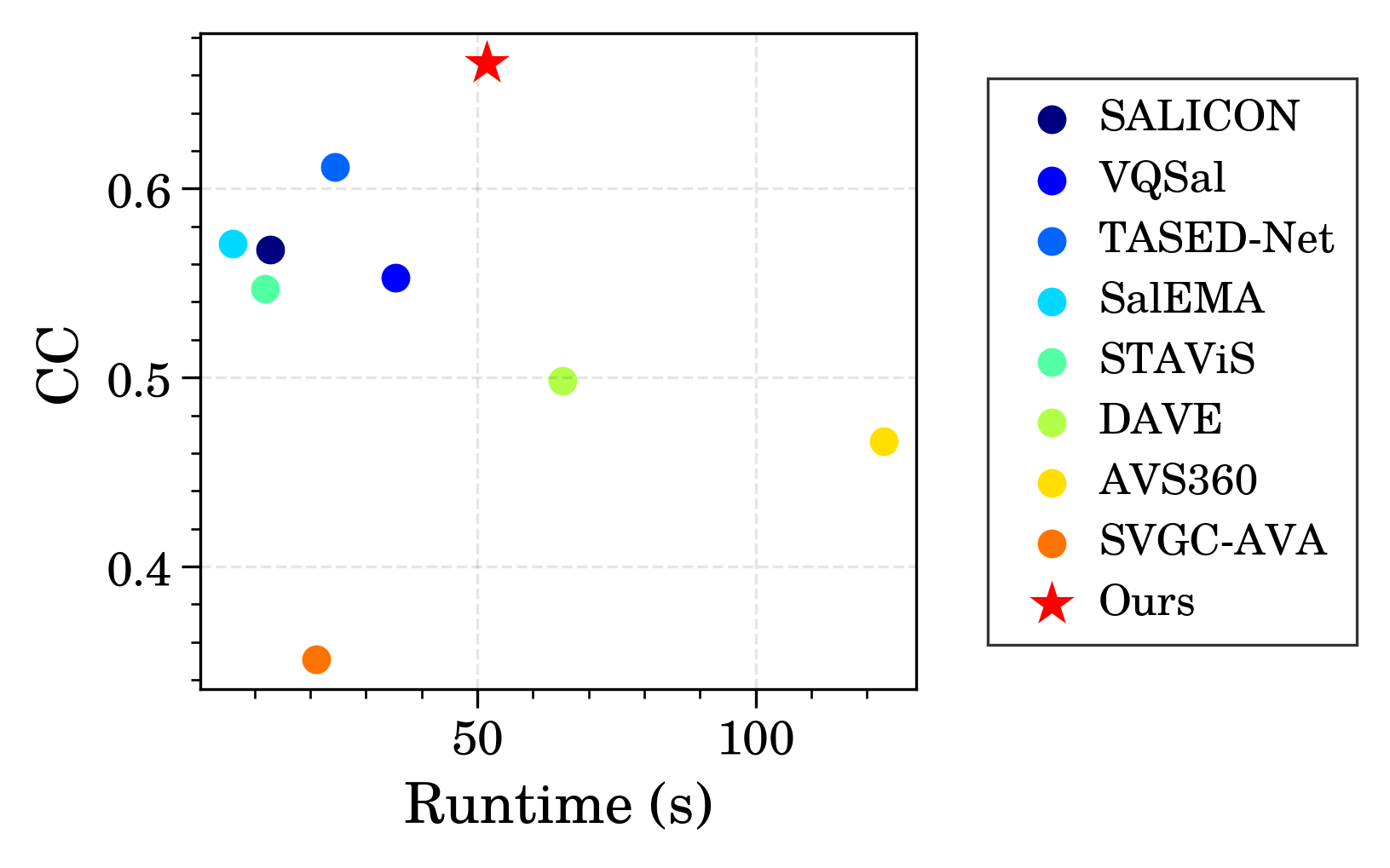}
\caption{Scatter plots of the average CC versus the running time for each saliency prediction model.}
\label{fig:runtime}
\vspace{-0.3cm}
\end{figure}

Considering the significance of computational efficiency in saliency prediction, we evaluate the computational efficiency of different deep learning models. We conduct runtime tests on a Linux server with an 80-core Intel(R) Xeon(R) Gold 5218R CPU @ 2.10GHz and an NVIDIA GeForce RTX 4090 GPU. We report the average running time for generating saliency maps at a resolution of $720\times360$ for a single video with 450 frames in AVS-ODV in Fig. \ref{fig:runtime}. It shows that OmniAVS can achieve the best prediction performance while maintaining relatively high computational efficiency.

\section{Conclusion and Future Works}\label{conclusion}
In this research, we have comprehensively investigated audio-visual attention in omnidirectional videos. We introduce AVS-ODV, currently the largest database for audio-visual saliency in ODVs, which includes 162 videos across various scenarios with user attention data under three audio settings. Through rigorous qualitative and quantitative analyses, we conclude that audio, particularly spatial audio, significantly influences visual attention in ODVs. This influence is most notable in scenes with multiple visual salient objects but only one sound source. Then we propose a novel model, OmniAVS, which effectively integrates visual and audio cues. Experimental results reveal that OmniAVS not only excels in predicting audio-visual saliency in omnidirectional contexts, but also demonstrates robust performance in saliency prediction for both visual-only ODVs and traditional flat videos.

While this study provides significant insights, there are several avenues for future work. 

The videos in the AVS-ODV database are self-shot and consist primarily of pure user-generated content (UGC), with minimal consideration given to background sounds or narrations. As the omnidirectional video is still an emerging media form, even on platforms like YouTube, ODVs with background audio (such as video narrations) are rare. However, in video media, it is common to use background audio that is not directly related to the visual objects, such as video narrations. Such audio can still influence visual attention through the semantic information it conveys.
In future work, we plan to expand the database to cover these scenarios. This will provide further insights into how non-object-related audio influences attention, enhancing our understanding of the influence of background audio on attention in immersive video environments, and contributing to more refined models for audio-visual saliency prediction.

Another direction for future work is to investigate the impact of depth on visual attention in ODVs with spatial audio. Incorporating depth information would better align with the immersive nature of these experiences, potentially enhancing model accuracy and realism.





\bibliographystyle{IEEEtran}
\bibliography{reference.bib}

\newpage

 




\vfill

\end{document}